%% file: main.tex
\documentclass[11pt]{article}

\usepackage[preprint]{acl}

\usepackage{times}
\usepackage{latexsym}

\usepackage[T1]{fontenc}
\usepackage[utf8]{inputenc}

\usepackage{microtype}

\usepackage{inconsolata}
\usepackage{graphicx}
\usepackage{multirow}
\usepackage{url}
\usepackage{kotex}
\usepackage{float}
\usepackage{booktabs}
\usepackage{tabularx}
\usepackage{makecell}
\usepackage{dblfloatfix}  
\usepackage{listings}
\usepackage[most]{tcolorbox}

\usepackage{graphicx}
\usepackage{booktabs}
\usepackage{algorithm}
\usepackage{algpseudocode}

\usepackage{bbm}
\usepackage{hyperref}
\usepackage{cleveref}
\crefname{figure}{Figure}{Figures}     
\Crefname{figure}{Figure}{Figures}  
\usepackage{amsfonts}
\usepackage{pifont}
\usepackage{xcolor}

\usepackage{tikz}
\newcommand*\circled[1]{\tikz[baseline=(char.base)]{\node[shape=circle,draw,inner sep=1pt] (char) {#1};}}
\newcommand{\smallcircled}[1]{\scalebox{1.0}{\circled{\scriptsize #1}}}
\title{Dependency-Aware Revocable Decoding for \\ Efficient Diffusion Large Language Model Inference}

\author{
 \textbf{Wooje Park\textsuperscript{1*}},
 \textbf{Insu Lee\textsuperscript{1*}},
 \textbf{Minyoung Noh\textsuperscript{1}},
 \textbf{Jaeyun Jang\textsuperscript{1}},
\\
 \textbf{Sungmin Lee\textsuperscript{1}},
 \textbf{Kyuhong Shim\textsuperscript{2}},
 \textbf{Byonghyo Shim\textsuperscript{1}}
\\
 \textsuperscript{1}Seoul National University,
 \textsuperscript{2}Sungkyunkwan University
\\
  \texttt{\{wjpark, islee, mynoh, jyjang, sungminlee, bshim\}@islab.snu.ac.kr}, 
  \texttt{khshim@skku.edu} \\
}

\begin{document}
\maketitle

\renewcommand{\thefootnote}{}
\footnotetext{
\textsuperscript{$*$} Co-first authors.
% \textsuperscript{$\dagger$} Co-corresponding authors.
}

\renewcommand{\thefootnote}{\arabic{footnote}} 
\input{sections/1_intro}

\input{sections/2_related_work}
\input{sections/3_method}

\input{sections/4_experiments}

\input{sections/5_analysis}

\input{sections/6_conclusion}

\section*{Limitations}
Our results show that DARD consistently improves the speed--quality trade-off over recent revocable decoding baselines across diverse benchmarks. 
At the same time, the absolute performance gain remains moderate. 
This is partly because our goal is not to alter the model distribution, but to improve sequence-level consistency by reusing the original dLLM prediction behavior.
In this sense, DARD is designed to enhance decoding reliability while preserving the base model's generation quality.
In addition, DARD introduces additional per-step computation through the augmented shadow sequence and state-specific attention masks. 
This overhead is small in practice, especially with block decoding, but more optimized implementations could further reduce the runtime cost.

% \section*{Acknowledgments}

% This document has been adapted
% by Steven Bethard, Ryan Cotterell and Rui Yan
% from the instructions for earlier ACL and NAACL proceedings, including those for
% ACL 2019 by Douwe Kiela and Ivan Vuli\'{c},
% NAACL 2019 by Stephanie Lukin and Alla Roskovskaya,
% ACL 2018 by Shay Cohen, Kevin Gimpel, and Wei Lu,
% NAACL 2018 by Margaret Mitchell and Stephanie Lukin,
% Bib\TeX{} suggestions for (NA)ACL 2017/2018 from Jason Eisner,
% ACL 2017 by Dan Gildea and Min-Yen Kan,
% NAACL 2017 by Margaret Mitchell,
% ACL 2012 by Maggie Li and Michael White,
% ACL 2010 by Jing-Shin Chang and Philipp Koehn,
% ACL 2008 by Johanna D. Moore, Simone Teufel, James Allan, and Sadaoki Furui,
% ACL 2005 by Hwee Tou Ng and Kemal Oflazer,
% ACL 2002 by Eugene Charniak and Dekang Lin,
% and earlier ACL and EACL formats written by several people, including
% John Chen, Henry S. Thompson and Donald Walker.
% Additional elements were taken from the formatting instructions of the \emph{International Joint Conference on Artificial Intelligence} and the \emph{Conference on Computer Vision and Pattern Recognition}.

\bibliography{custom}

% \newpage{
%     \bibliography{custom}
% }

% \clearpage 
% \input{summary_of_resub}

\clearpage
\appendix
\input{appendix/appendix}

\end{document}

%% file: sections/1_intro.tex
\begin{abstract}
Diffusion large language models (dLLMs) offer a promising alternative to autoregressive generation by decoding multiple tokens in parallel through iterative denoising.
However, increasing decoding parallelism often degrades generation quality, as early errors can contaminate later contexts.
Revocable decoding mitigates this issue by re-evaluating decoded tokens and remasking unreliable ones, but existing methods overlook that unreliable tokens may also corrupt the verification context itself.
We identify this failure mode and propose Dependency-Aware Revocable Decoding (DARD), a training-free framework that separates tokens into masked, candidate, and unmasked states.
DARD verifies candidate tokens using a selective context that excludes less reliable tokens and adaptively regulates their influence on subsequent decoding.
Experiments across 12 textual and multimodal benchmarks on 3 open-source dLLMs show that DARD consistently improves the speed-quality Pareto frontier over recent revocable decoding methods, achieving a 2.71$\times$ speedup and a 4.35-point CIDEr score gain over Saber on Flickr30K.
\end{abstract}

\section{Introduction}

In recent years, diffusion large language models (dLLMs) have gained attention as a compelling alternative to autoregressive large language models (AR-LLMs), which have long served as the de facto standard for natural language generation~\cite{achiam2023gpt4,yang2025qwen3,liu2024deepseekv3,olmo2025olmo3,team2024gemma2}.
In contrast to AR-LLMs generating text tokens one at a time from left to right, dLLMs generate text by iteratively denoising masked sequences~\cite{sahoo2024simple,lou2023SEDD,ou2025RADD,ye2025dream,zhu2025llada1.5}.
At each denoising step, dLLMs predict token distributions for all masked positions and then selectively unmask a subset of promising positions exhibiting high confidence scores (i.e., the top-1 probabilities)~\cite{nie2026llada}.
In doing so, multiple token positions can be decoded in parallel, speeding up the inference process.

\input{figures/1_intro}

A well-known drawback of this approach is that the quality of generated text degrades sharply when the number of tokens decoded at each step increases~\cite{kang2025parallelbench}.
\input{figures/1_intro_example}
A key reason is that each masked position is predicted using only the context available at that step, without referencing the tokens predicted for other positions in the same step.
As a result, tokens predicted from such limited context may turn out to be inconsistent with the context revealed in later steps.
Once committed, these erroneous tokens keep providing misleading context for future predictions, propagating errors throughout the generation process.

To address the problem, recent studies have explored revocable decoding strategies for dLLMs. 
Instead of treating decoded tokens as fixed ones after unmasking, these methods re-evaluate them using the updated context and then remask unreliable ones among them.
For example, Saber~\cite{dong2025saber} identifies suspicious tokens by tracking confidence drops across decoding steps, while WINO~\cite{hong2025wino} uses an auxiliary verification path to re-evaluate decoded tokens and remask those with low confidence.
These methods demonstrate improvements in the speed-quality trade-off in dLLMs to some extent.

Notwithstanding these benefits, existing revocable methods leave an important issue unresolved: erroneous tokens may also be used as context when verifying other tokens, which can undermine the reliability of the verification process.
To illustrate the problem, we consider a scenario in which two masked positions are decoded in parallel (see Figure~\ref{fig:intro_example})~\cite{song2025intro_example}.
For the prompt ``The city of \_ \_ is on the Southern California coast'', valid completions include ``Los Angeles'' and ``San Diego''.
Since the two positions are predicted within the same decoding step, the model may incorrectly generate ``Los Diego'', where each token is locally plausible but the pair is invalid.
Existing methods may fail to correct such errors because each token is verified independently while the other decoded token remains visible as context.
For example, the model may re-evaluate ``Los'' as ``San'' with ``Diego'' as context, and ``Diego'' as ``Angeles'' with ``Los'' as context.
These conflicting verification signals may lead the model to regard both positions as unreliable and remask them unnecessarily.

An aim of this paper is to propose a revocable decoding framework that mitigates verification errors caused by unreliable tokens.
The core idea of the proposed method, dubbed Dependency-Aware Revocable Decoding (DARD), is to verify each token under a selective context that excludes unreliable tokens.
To realize this idea, we introduce three token states: masked ($\mathcal{M}$) for undecoded tokens, candidate ($\mathcal{C}$) for decoded but uncertain tokens, and unmasked ($\mathcal{U}$) for decoded tokens with high confidence (see Figure~\ref{fig:1_intro}).
By further distinguishing decoded tokens according to their reliability, DARD avoids treating uncertain tokens as fully reliable context during verification.
Specifically, each $\mathcal{C}$ token attends only to more reliable tokens, i.e., all $\mathcal{U}$ tokens and higher-confidence $\mathcal{C}$ tokens.
This design mimics confidence-ordered multi-step decoding, where higher-confidence candidates are treated as if they were decoded earlier and used as context for verification.
Additionally, we leverage the verification results to estimate the reliability of the $\mathcal{C}$ set and adaptively control how strongly $\mathcal{M}$ tokens rely on $\mathcal{C}$ tokens during decoding.

Extensive experiments across 12 textual and multimodal benchmarks on 3 open-source dLLMs show that DARD consistently improves the speed-quality Pareto frontier over recent revocable decoding methods.
In particular, on Flickr30K~\cite{young2014flickr30k}, DARD achieves a 2.71$\times$ speedup while improving CIDEr score by 4.35-points over Saber~\cite{dong2025saber}.
These results demonstrate that a reliable verification context is a key ingredient for effective revocable decoding in dLLMs.

\noindent In summary, our contributions are as follows:
\begin{itemize}
    \item In this work, we identify a previously overlooked failure mode in revocable decoding frameworks: unreliable tokens can contaminate the verification context, leading to persistent errors or unnecessary remasking.

    \item To mitigate verification errors, we propose DARD, a training-free revocable decoding framework that uses three token states to selectively control contextual dependencies among decoded tokens during verification.

    \item We show that DARD improves the speed-quality Pareto frontier across 12 benchmarks and 3 representative open-source dLLMs, with additional ablations and case studies further validating its effectiveness and efficiency.
\end{itemize}

%% file: figures/1_intro.tex
\begin{figure}[t!]
  \centering
    \includegraphics[width=\linewidth]{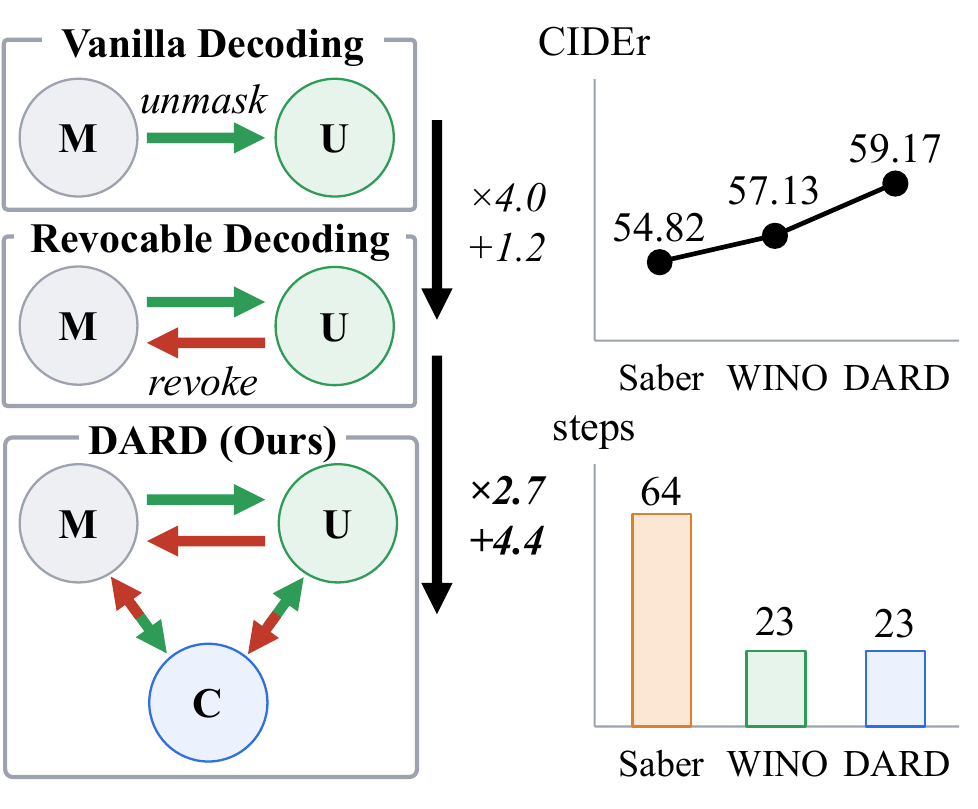}
  \caption{
  Left: Comparison of decoding schemes.
  Vanilla dLLM decoding fixes tokens once unmasked, while revocable decoding allows them to be remasked.
  DARD introduces an intermediate $\mathcal{C}$ state to place unstable decoded tokens and regulate their contextual influence during verification.
  Right: By selectively using context, DARD mitigates verification errors arising from context instability, thereby reducing decoding steps while improving performance.
  }
  \label{fig:1_intro}
\end{figure}

%% file: figures/1_intro_example.tex
\begin{figure*}[t!]
  \centering
  \includegraphics[width=\textwidth]{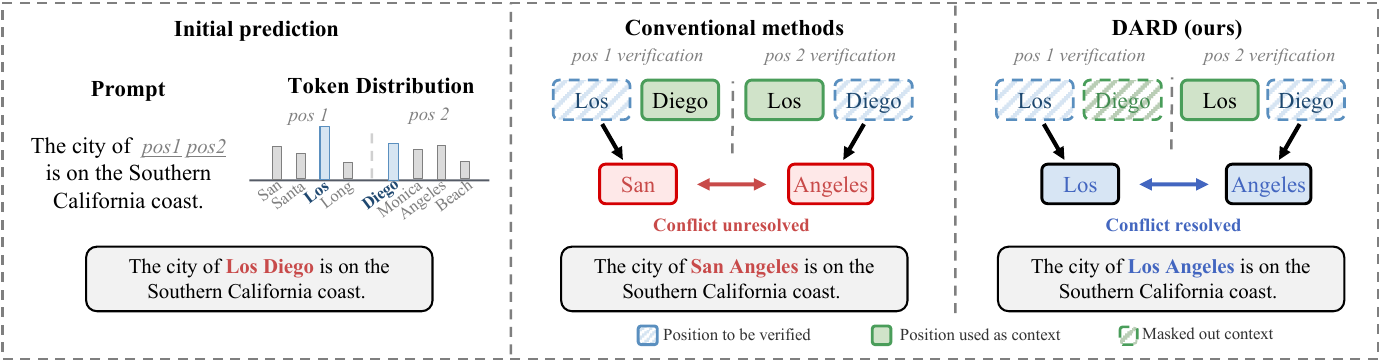}% or \columnwidth
  \caption{
  Left: Parallel decoding predicts multiple token positions simultaneously, making it difficult to account for dependencies among tokens and potentially producing inconsistent predictions.
  Center: Conventional revocable methods verify tokens based on all decoded tokens, so inconsistent tokens may mislead verification.
  Right: DARD performs dependency-aware verification, where each token attends only to more reliable context tokens.
}
  \label{fig:intro_example}
\end{figure*}

%% file: sections/2_related_work.tex
\section{Related Work}

\input{figures/3_method_main}

\subsection{Diffusion Large Language Models}

Diffusion large language models (dLLMs) generate text by iteratively denoising masked sequences, which allows tokens to be generated in a flexible order and decoded in parallel.
These capabilities have attracted growing interest and have been explored in commercial systems including Mercury, Gemini Diffusion, and Seed Diffusion~\cite{khanna2025mercury,google2025geminidiffusion,song2025seeddiffusion}.
This interest has also extended to the open-source community.
The LLaDA series has demonstrated the viability of large-scale native diffusion architectures trained from scratch, with later variants improving alignment and architectural efficiency~\cite{nie2026llada,zhu2025llada1.5,zhu2025llada-moe}.
In parallel, autoregressive-to-diffusion approaches, such as DiffuLLaMA and Dream, adapt pretrained AR-LLMs into dLLMs to reduce training cost while preserving the parallel decoding capability of diffusion models~\cite{gong2025diffullama,ye2025dream,fu2025efficientdlm,bie2025llada2.0}. 
Recent work has also extended dLLMs beyond language-only modeling to multimodal generation and understanding~\cite{yang2026mmada,li2026lavida,you2025llada-v}.
Despite these advances, dLLMs still suffer from a speed-quality trade-off during inference, motivating more effective decoding strategies.

\subsection{dLLM Acceleration Techniques}
A growing body of work has sought to accelerate dLLM inference without sacrificing generation quality.
One line of work reduces per-step computation by adapting attention Key-Value (KV) caching to dLLMs.
This adaptation is nontrivial because, in dLLMs, bidirectional attention causes KV states to change across denoising steps.
Prior works address this issue through block-wise generation, approximate KV reuse, delayed caching, or pruning-based strategies~\cite{arriola2025block-diffusion,wu2025fast-dllm,liu2025dllm-cache,hu2025flashdlm,ma2026dkv-cache,song2026sparse-dllm}.
Another line of work reducing the number of denoising steps is to decode more tokens at each step.
Some methods exploit model-internal signals, such as confidence, entropy, or probability margins, to determine which tokens can be safely committed~\cite{wu2025fast-dllm,ben2026EBsampler,li2025Prophet}.
Others learn decoding policies or improve token certainty through additional training~\cite{bao2025Learn2PD,chen2025dparallel}.
More recently, revocable decoding methods make decoding decisions revisable, aggressively decoding multiple tokens in parallel and remasking suspicious tokens after verification~\cite{hong2025wino,dong2025saber}.
Further discussion of concurrent work is provided in Appendix~\ref{comp_with_con_work}.

\input{figures/3_method_sub}

%% file: figures/3_method_main.tex
\begin{figure*}[t!]
  \centering
  \includegraphics[width=\textwidth]{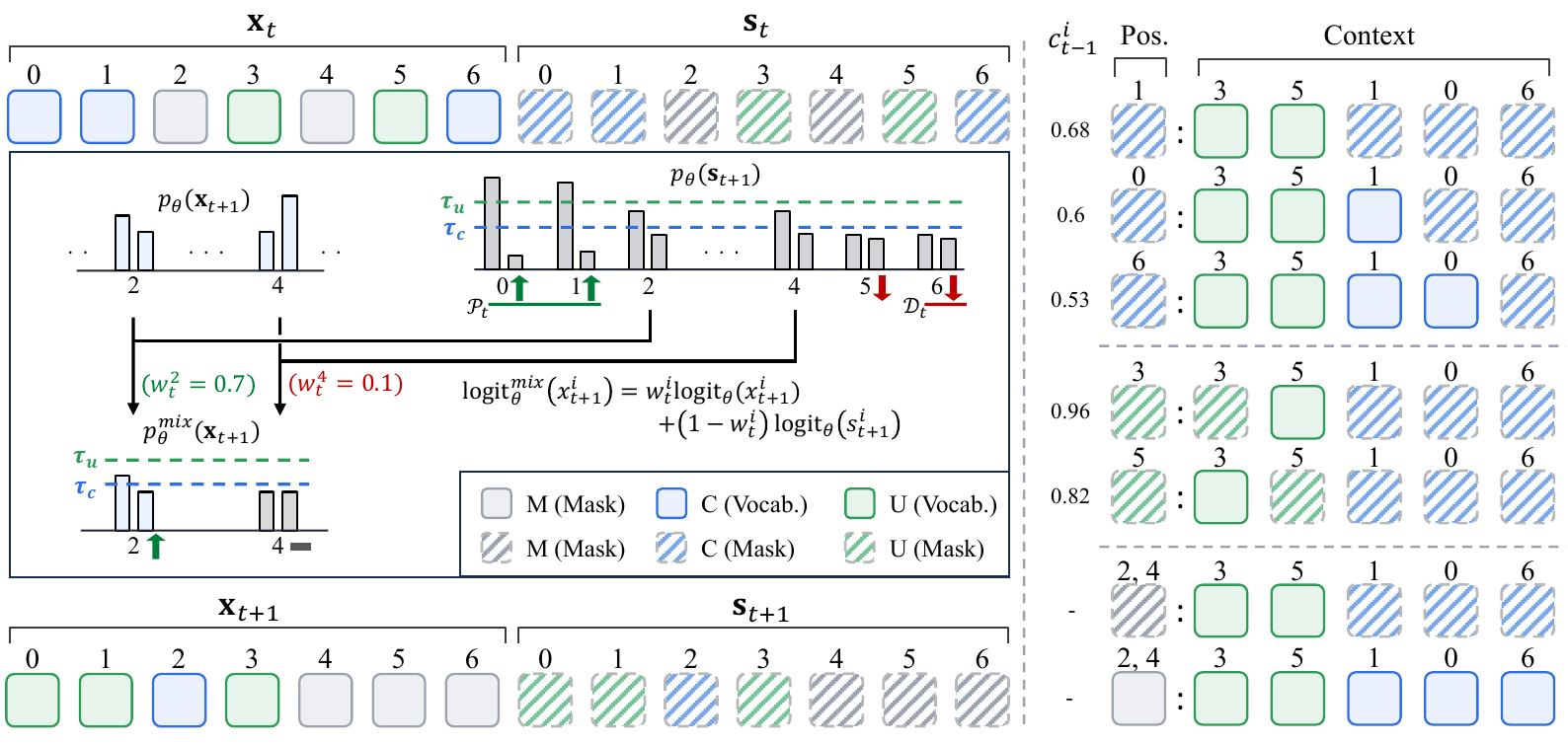}% or \columnwidth
  \caption{
Overview of DARD.
DARD verifies tokens using state-specific contexts defined according to token reliability and updates their states at the next decoding step.
Based on the verification outcomes, DARD estimates the reliability of $\mathcal{C}$ tokens and adaptively controls their contribution to $\mathcal{M}$ token prediction.
}
  \label{fig:method}
\end{figure*}

%% file: figures/3_method_sub.tex
\begin{figure}[t!]
  \centering
    \includegraphics[width=\linewidth]{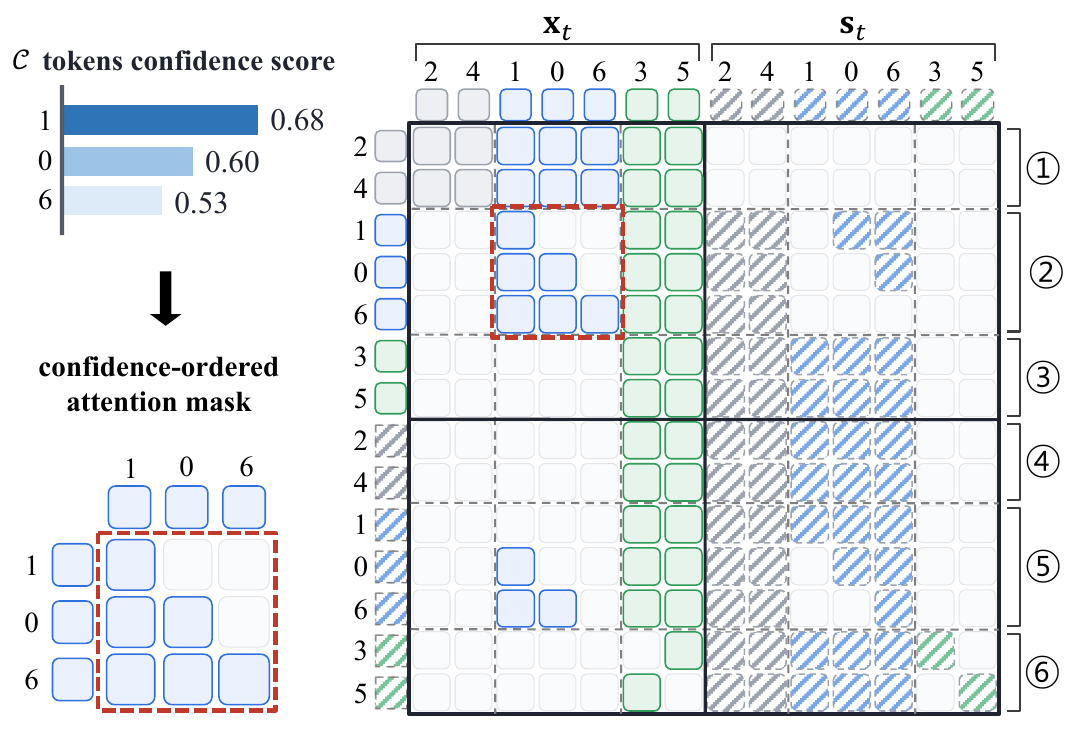}
  \caption{Attention mask design of DARD.
  Rows correspond to queries, and columns correspond to keys.
  DARD uses state-specific attention masks to construct different verification contexts for different token states.
  }
  \label{fig:3_method_sub}
\end{figure}

%% file: sections/3_method.tex
\section{Dependency-Aware Revocable Decoding}
\label{sec:method}

\subsection{Preliminaries}
\label{sec:preliminaries}
\paragraph{Inference process in dLLMs.}

We consider absorbing dLLMs~\cite{nie2026llada} parameterized by $\theta$, which generate the target sequence $\mathbf{x}_T$ by progressively denoising partially masked sequences $\mathbf{x}_t = (x_t^0, \ldots, x_t^{L-1})$ over discrete timesteps $t \in \{0,\dots ,T-1\}$.
The decoding process starts from $\mathbf{x}_0$, where $x_0^i = \texttt{[Mask]}$ for all $i \in \{0,\ldots,L-1\}$.
At step $t$, the model estimates a categorical distribution $p_\theta$ over the vocabulary $\mathcal{V}$ for each masked position $i \in \mathcal{M}_t$, where $\mathcal{M}_t:= \{i \mid x_t^i = \texttt{[Mask]}\}$. 
We denote the resulting prediction and confidence by
\begin{equation}
    \begin{aligned}
        \hat{x}_{t}^i
        =\:&
        \arg\max_{v \in \mathcal{V}}
        p_\theta(x_{t+1}^i = v \mid \mathbf{x}_t), \\
        c_t^i
        &=
        p_\theta(x_{t+1}^i = \hat{x}_{t}^i \mid \mathbf{x}_t).
    \end{aligned}
\end{equation}
Based on $c_t^i$, we select a subset of masked positions and replace the token at each position $i$ with $\hat{x}_t^i$ to obtain $\mathbf{x}_{t+1}$.

\paragraph{Parallelized inference over distinct decoding contexts.}
In this work, we introduce a length-$L$ shadow sequence $\mathbf{s}_t=(s_t^0,\ldots,s_t^{L-1})$, which is fully masked throughout the decoding process~\cite{hong2025wino}.
To parallelize token prediction and verification, we concatenate $\mathbf{x}_t$ with $\mathbf{s}_t$ to form the augmented input $\tilde{\mathbf{x}}_t=(\mathbf{x}_t;\mathbf{s}_t)$.
For each position $i$, $x_t^i$ and $s_t^i$ share the same positional embedding, and the model produces two predictions, $p_\theta(x_{t+1}^i\mid\tilde{\mathbf{x}}_t)$ and $p_\theta(s_{t+1}^i\mid\tilde{\mathbf{x}}_t)$, respectively.

To condition these predictions on distinct contexts, we apply a binary attention mask $M\in\{0,1\}^{2L\times 2L}$ over $\tilde{\mathbf{x}}_t$, where $M_{ij}=0$ indicates that attention from query position $i$ to key position $j$ is blocked.
Let $\bar{j}:=L+j$ denote the shadow position in $\tilde{\mathbf{x}}_t$ corresponding to original position $j$.
For each query position $i$, the mask satisfies
\begin{equation}
M_{ij} + M_{i\bar{j}} = 1.
\end{equation}
This constraint ensures that queries in both the original and shadow sequences attend to exactly one of $x_t^j$ and $s_t^j$ at every position $j$.
The original and shadow queries can therefore be conditioned on different contexts and computed in parallel within a single forward pass.

\subsection{State-Specific Prediction and Verification}
\paragraph{Overall decoding procedure.}
Our decoding procedure is designed to fully leverage information from all decoded tokens while reducing the risk of using incorrectly decoded tokens as conditioning context.
Accordingly, we introduce a candidate ($\mathcal{C}$) state between the conventional masked ($\mathcal{M}$) and unmasked ($\mathcal{U}$) states so that the model can verify uncertain tokens separately and selectively use them as conditioning context.
The decoding process starts with all tokens assigned to the $\mathcal{M}$ state. 
At each step $t$, the model determines each token’s state by comparing its confidence $c_t^i$ with the thresholds $\tau_c$ and $\tau_u$, as specified in Equation~\ref{eq:state_transition}.
Based on these states, the model verifies $\mathcal{C}$ and $\mathcal{U}$ tokens using predictions from the shadow sequence.
For $\mathcal{M}$ tokens, the model combines predictions from the original and shadow sequences.
The overall decoding procedure of our method is illustrated in Figure~\ref{fig:method}.

\paragraph{Verification of $\mathcal{U}$ tokens with reliable context.}
In our framework, the $\mathcal{U}$ state represents decoded token positions that are currently considered reliable, where $\mathcal{U}_t:= \{i \mid x_t^i \neq \texttt{[Mask]},\; \tau_{\mathrm{u}} < c_{t-1}^i\}$.
During the verification of $\mathcal{U}_t$ tokens, we restrict their conditioning context to tokens currently considered reliable.
Specifically, $\mathcal{U}_t$-token queries attend to the keys of $\mathcal{U}_t$ tokens in the original sequence $\mathbf{x}_t$.
These queries attend to the keys of $\mathcal{M}_t$ and $\mathcal{C}_t$ tokens in the shadow sequence $\mathbf{s}_t$ (query blocks \smallcircled{3} and \smallcircled{6} in Figure~\ref{fig:3_method_sub}).
For each query of $s_t^i$, we block attention to the corresponding key of $x_t^i$ to prevent information leakage from the decoded token at the same position.
Although verification relies only on $p_\theta(s_{t+1}^i\mid\tilde{\mathbf{x}}_t)$, we apply the same attention pattern to the original sequence queries because their intermediate representations serve as keys in subsequent layers.
This design excludes relatively uncertain $\mathcal{C}_t$ tokens from the conditioning context, thereby preventing them from contaminating the verification results.

\paragraph{Verification of $\mathcal{C}$ tokens with confidence-ordered context.}

We define $\mathcal{C}_t \!:= \!\{i \!\mid\! x_t^i \neq \texttt{[Mask]},\; \tau_{\mathrm{c}}\!< \!c_{t-1}^i\! \leq\! \tau_{\mathrm{u}}\}$ as the set of decoded positions that require further verification before serving as reliable context.
Our use of confidence to order tokens in $\mathcal{C}_t$ is motivated by prior work that treats confidence as a proxy for token reliability during decoding~\cite{kim2025train, wu2025fast-dllm}.
Since tokens in $\mathcal{C}_t$ have relatively low model confidence, using them indiscriminately as context for other tokens may compromise verification reliability.
To avoid this contamination, we apply a confidence-ordered attention mask that blocks information flow from lower-confidence tokens to higher-confidence tokens:
\begin{equation}
    M_{ij}
    =
    \begin{cases}
        1, & \text{if } c_{t-1}^i \leq c_{t-1}^j, \\
        0, & \text{if } c_{t-1}^i > c_{t-1}^j,
    \end{cases}
    \quad i,j \in \mathcal{C}_t.
\end{equation}
In addition to attending to the keys of $\mathcal{U}_t$ tokens in $\mathbf{x}_t$, $\mathcal{C}_t$-token queries in both $\mathbf{x}_t$ and $\mathbf{s}_t$ attend only to the keys of higher-confidence $\mathcal{C}_t$ tokens in $\mathbf{x}_t$ (query blocks \smallcircled{2} and \smallcircled{5} in
Figure~\ref{fig:3_method_sub}).

This masking rule can also be interpreted in an autoregressive-like manner, where inter-token dependencies are ordered by confidence rather than by left-to-right position (see Appendix~\ref{dependency_details} for
details):
\begin{equation}
% \nonumber
    p_\theta\!
    \bigl(
    s_{t+1}^{\pi_t(k)}\!
    \!\mid\!
    \tilde{\mathbf{x}}_{t};\! M
    \bigr)\!
    \approx \!
    p_\theta
    \bigl(
    s_{t+1}^{\pi_t(k)}
    \!\!\mid\!
    \hat{x}_{t}^{\pi_t(1:k-1)}\!,
    \hat{x}_t^{\mathcal{U}_t}
    \bigr),\!\!
\end{equation}
for $k=1,\ldots,|\mathcal{C}_t|$, where $\pi_t(k)$ denotes the $k$-th position in $\mathcal{C}_t$ under descending confidence order.
The underlying motivation is to approximate confidence-ordered multi-step decoding, in which higher-confidence $\mathcal{C}_t$ tokens are decoded earlier and subsequently provide context for lower-confidence tokens.
Appendix~\ref{app:conf_order} shows that this ordering closely aligns with the confidence-based decoding trajectory.

\paragraph{Adaptive prediction of $\mathcal{M}$ tokens based on verification results.}

For each masked position $i \in \mathcal{M}_t$ (i.e., $c_{t-1}^i \leq \tau_{\mathrm{c}}$), the model uses predictions from both $\mathbf{x}_t$ and $\mathbf{s}_t$.
We apply different attention masks for the two predictions so that they complement each other in both cases: when $\mathcal{C}_t$ tokens serve as reliable context and when they serve as misleading context.
Specifically, $\mathcal{M}_t$-token queries in  $\mathbf{x}_t$ attend to the keys of $\mathcal{C}_t$ and $\mathcal{U}_t$ tokens in $\mathbf{x}_t$.
The corresponding queries in $\mathbf{s}_t$ attend to the keys of $\mathcal{U}_t$ tokens in $\mathbf{x}_t$ and of $\mathcal{C}_t$ tokens in $\mathbf{s}_t$, regarding $\mathcal{C}_t$ as masked tokens (query blocks \smallcircled{1} and \smallcircled{4} in Figure~\ref{fig:3_method_sub}).

We adaptively combine the two predictions based on the verification results of tokens in $\mathcal{C}_t$.
Let $\mathcal{P}_t$ denote the positions in $\mathcal{C}_t$ that are promoted to $\mathcal{U}_{t+1}$, and let $\mathcal{D}_t$ denote those that are demoted to $\mathcal{M}_{t+1}$. 
For each position $i\in\mathcal{M}_t$, we compute two distance-weighted scores:
\begin{equation}
    P_t^i = \sum_{j \in \mathcal{P}_t} K(i,j),
    \quad
    D_t^i = \sum_{j \in \mathcal{D}_t} K(i,j),
\end{equation}
where $K(i,j) = \lambda^{|i-j|}$ is a geometric kernel function with $0 < \lambda < 1$, so that the influence of each verification outcome decreases with its distance from position $i$~\cite{ye2025dream}. 
The guidance weight $w_t^i$ for each position $i$ is then defined as
\begin{equation}
\label{eq:logit_mixing}
    w_t^i =
    \frac{P_t^i + p_0}
    {P_t^i + D_t^i + p_0},
\end{equation}
where $p_0$ is the prior constant stabilizing the weight when only a few state changes occur~\cite{manning2008introduction}.
Thus, $w_t^i$ increases when nearby $\mathcal{C}_t$ tokens are predominantly promoted rather than demoted, thereby assigning greater weight to the prediction conditioned on $\mathcal{C}_t$.
Using this weight, we combine the two predictions for each $i\in\mathcal{M}_t$ as follows:
\begin{equation}
    \begin{aligned}
        &\text{logit}_\theta^{mix}
        (x_{t+1}^{i}\mid \tilde{\mathbf{x}}_{t};M) \\
        &\qquad=w_t^i\,\text{logit}_\theta
        (x_{t+1}^{i}\mid \tilde{\mathbf{x}}_{t};M)\\
        &\quad\qquad + (1-w_t^i)\,
        \text{logit}_\theta
        (s_{t+1}^{{i}}\mid \tilde{\mathbf{x}}_{t};M),
    \end{aligned}
\end{equation}
where $\text{logit}_\theta(\cdot)$ denotes the scores before normalization into $p_\theta(\cdot)$.

\input{figures/4_experiment_llm}

\paragraph{State transition.}
We formulate decoding as a three-state transition system over token positions.
Given the confidence score $c_t^i$, the state of each position $i$ is updated according to the following rule: 
\begin{equation}
i
\quad \rightarrow \quad
\begin{cases}
 \mathcal{M}_{t+1}, & c_t^i \leq \tau_{\mathrm{c}}, \\
\mathcal{C}_{t+1}, & \tau_{\mathrm{c}} < c_t^i \leq \tau_{\mathrm{u}}, \\
\mathcal{U}_{t+1}, & \tau_{\mathrm{u}} < c_t^i,
\end{cases}
\label{eq:state_transition}
\end{equation}
where $\tau_{\mathrm{c}} \leq \tau_{\mathrm{u}}$.
Based on this assignment, positions in $\mathcal{M}_{t+1}$ are set to
$\texttt{[Mask]}$, while positions in
$\mathcal{C}_{t+1}\cup\mathcal{U}_{t+1}$ are replaced with their corresponding
predictions $\hat{x}_t^i$.
The overall procedure is summarized in Algorithm~\ref{alg:state_transition}.

%% file: figures/4_experiment_llm.tex
\begin{figure*}[t!]
  \centering
  \includegraphics[width=\textwidth]{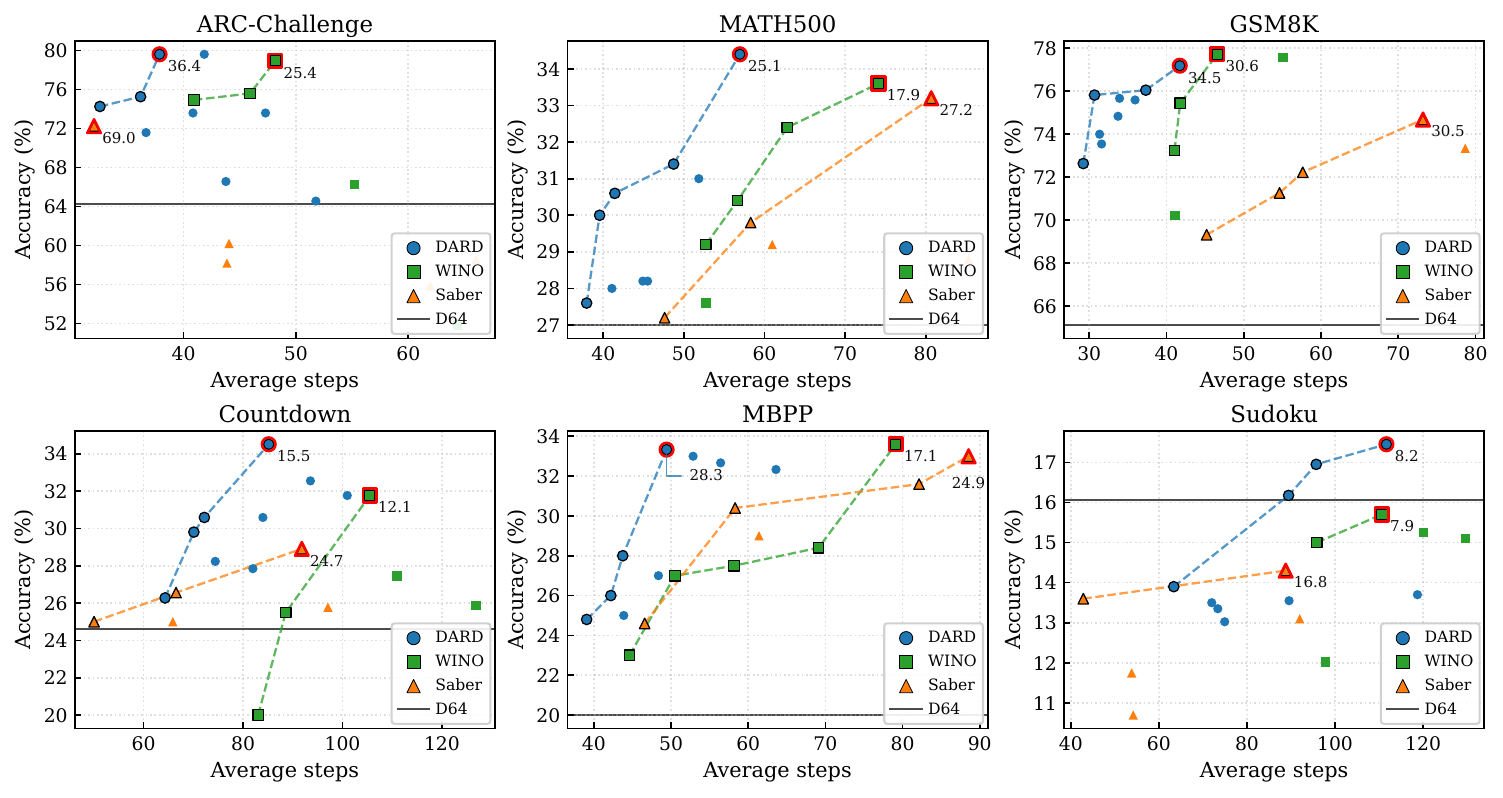}% or \columnwidth
  \caption{
LLaDA-8B-Instruct results on six language benchmarks.
  Each plot compares task performance against the average number of decoding steps for each method.
  Dashed curves indicate the Pareto frontier for each method.
  For each method, the best-performing configuration is highlighted with a red outline, and its TPS is annotated next to the point.
  Horizontal lines show the performance of standard LLaDA decoding with 64 steps.
  }
  \label{fig:experiment_llm}
\end{figure*}

%% file: sections/4_experiments.tex
\section{Experiments}

\subsection{Experiment Setup}

\paragraph{Evaluation Details.}
We evaluate the language benchmarks following the protocol of WINO~\cite{hong2025wino}, while all vision-language benchmarks are evaluated using the \texttt{lmms-eval} package~\cite{zhang2025lmms}.
All benchmarks are evaluated in a zero-shot setting, except for Sudoku~\cite{ye2025beyond_sudoku}, where we use a $4$-shot setting.
For task performance, we report accuracy for all benchmarks except Flickr30K~\cite{young2014flickr30k}, for which we report CIDEr.
For MBPP~\cite{austin2021program_mbpp}, we measure accuracy by pass@1 functional correctness, while for Sudoku, we use cell-level partial-credit accuracy.
We evaluate MathVista~\cite{lu2024mathvista} and MathVision~\cite{wang2024mathvision} using rule-based answer matching.
To evaluate inference efficiency, we report the average number of decoding steps per sample and the throughput measured in tokens per second (TPS).
Additional details on datasets, baselines, and implementation are provided in Appendix~\ref{app:experimental_details}.

\input{figures/4_experiment_lvlm}

\subsection{Main Results}

We report Pareto curves of task performance versus the average number of decoding steps.
Each point represents a decoding configuration, and points closer to the upper-left indicate better speed-quality trade-offs; dashed lines indicate the Pareto frontier of each method.

\paragraph{Language Benchmarks.}

Figure~\ref{fig:experiment_llm} shows the speed-quality trade-off on six language benchmarks using LLaDA-8B-Instruct~\cite{nie2026llada}.
Across the benchmarks, DARD generally achieves a more favorable Pareto frontier than WINO~\cite{hong2025wino} and Saber~\cite{dong2025saber}, attaining comparable or higher performance with fewer decoding steps.
Notably, DARD also performs well on task-specific benchmarks such as Countdown~\cite{gandhi2024stream_countdown} and Sudoku~\cite{ye2025beyond_sudoku}, suggesting that dependency-aware verification remains effective even under structured reasoning and strict-format requirements.
On MBPP~\cite{austin2021program_mbpp} and GSM8K~\cite{cobbe2021training_gsm8k}, although DARD shows a slight performance drop, it achieves similar performance with substantially fewer decoding steps.
Overall, these results demonstrate that DARD improves the speed-quality trade-off across diverse language tasks.
Additional results on LLaDA-1.5~\cite{zhu2025llada1.5} are provided in Appendix~\ref{app:llada15_results}.

\paragraph{Vision-Language Benchmarks.}

Figure~\ref{fig:experiment_lvlm} shows the speed-quality trade-off on six vision-language benchmarks using MMaDA-8B-MixCoT~\cite{yang2026mmada}.
DARD achieves a better Pareto frontier than recent revocable decoding methods across most benchmarks, showing higher performance with fewer decoding steps.
Notably, compared with Saber, DARD reduces the step count by more than $2\times$ while improving Flickr30K~\cite{young2014flickr30k} by over 4 CIDEr points and AI2D~\cite{kembhavi2016ai2d} by over 2 accuracy points.

%% file: figures/4_experiment_lvlm.tex
\begin{figure*}[t!]
  \centering
  \includegraphics[width=\textwidth]{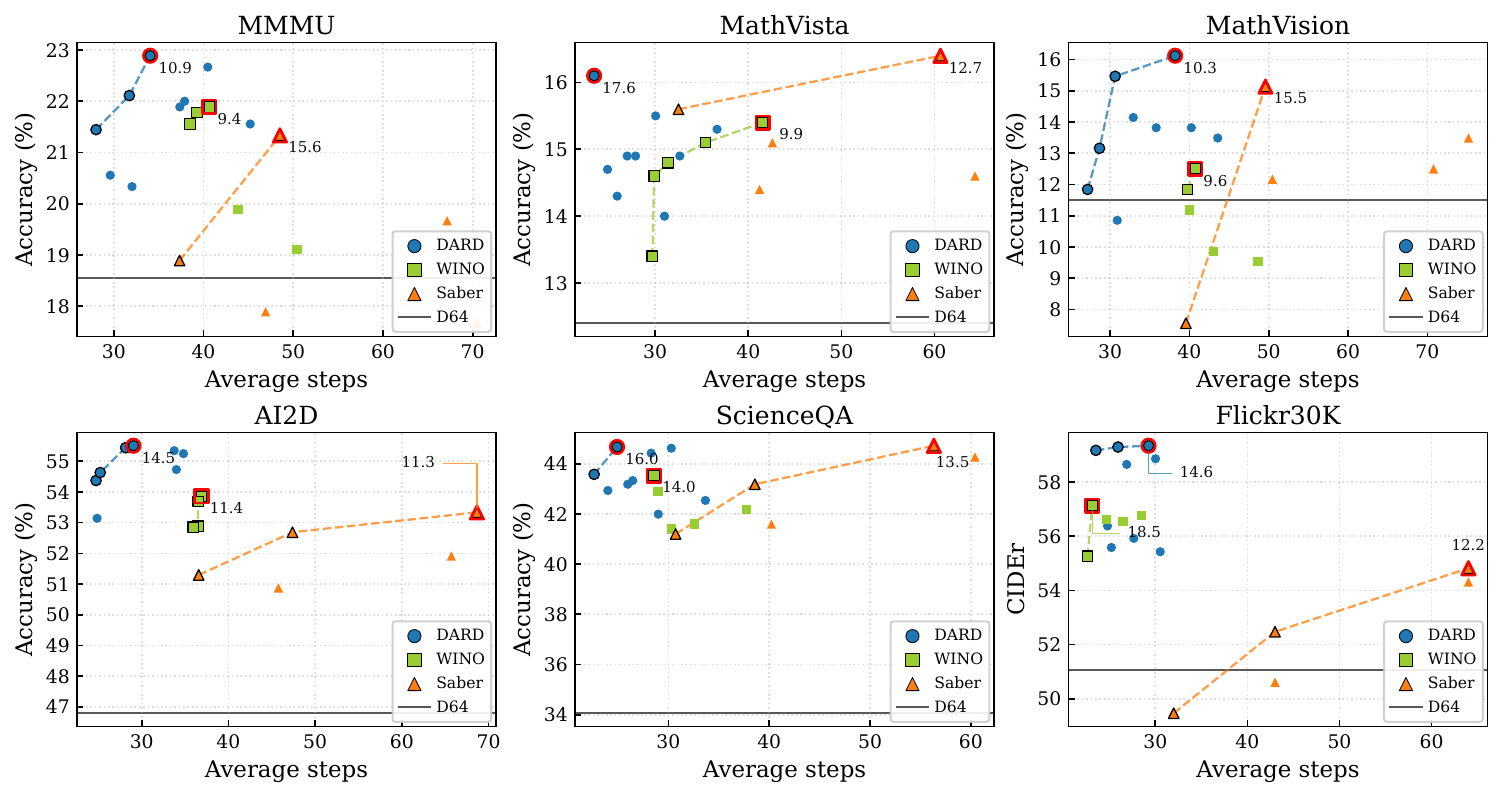}% or \columnwidth
  \caption{
  MMaDA-8B-MixCoT results on six vision-language benchmarks.
  Each plot compares task performance against the average number of decoding steps for each method.
  Dashed curves indicate the Pareto frontier for each method.
  For each method, the best-performing configuration is highlighted with a red outline, and its TPS is annotated next to the point.
  Horizontal lines show the performance of standard MMaDA decoding with 64 steps.
  }
  \label{fig:experiment_lvlm}
\end{figure*}

%% file: sections/5_analysis.tex
\section{Analysis}
Unless otherwise specified, all analyses are conducted on the MATH500 benchmark using the LLaDA-Instruct 8B model. 
We use the decoding configuration that achieves the highest accuracy in our main experiments as the default setting. 
\input{figures/5_qualitative}

\subsection{Analysis of DARD Design Choice}
\paragraph{Attention mask design.}
We analyze how the attention mask design affects the verification of tokens in $\mathcal{C}$ within DARD.
We first report the evaluation results using a bidirectional mask, where candidate tokens can attend to each other without any directional constraint.
In addition, we evaluate three alternative ordering strategies: two commonly used criteria in dLLMs decoding, namely entropy and margin~\cite{sahoo2024simple}, and the left-to-right (l2r) order used in AR-LLMs.

As shown in Table~\ref{tab:attention_mask_ablation}, applying a bidirectional mask reduces decoding steps by only 0.8, while causing a substantial accuracy drop of 3.2 percentage points.
This result suggests that allowing uncertain candidate tokens to freely interact with one another can make verification unstable, despite slightly improving decoding efficiency. 
In contrast, the ordering-based variants achieve comparable decoding efficiency, while our confidence-based ordering maintains substantially higher accuracy. 
This demonstrates the effectiveness of our proposed confidence-based ordering strategy.

\input{tables/attention_mask_ablation}

\paragraph{Adaptive logit mixing.}
We analyze the effect of adaptive logit mixing in DARD. 
The logits from $\mathbf{s}_t$ are computed from positions that do not attend to the context of the $\mathcal{C}$ tokens, whereas those from $\mathbf{x}_t$ fully leverage this context.
To examine how this contextual difference affects masked-token prediction, we evaluate three fixed weights, $w \in \{0.0, 0.5, 1.0\}$, where $w=0.0$ completely excludes the $\mathcal{C}$ token context, while $w=1.0$ fully incorporates it.

As shown in Table~\ref{tab:adaptive_logit_mixing}, fully leveraging the $\mathcal{C}$ token context ($w = 1.0$) achieves the fewest decoding steps, indicating that such contextual information helps accelerate subsequent decoding.
However, our logit mixing achieves higher accuracy with only a slight increase in decoding steps. 
This demonstrates the benefit of adaptively controlling the contribution of the $\mathcal{C}$ token context during mask prediction.

\input{tables/adaptive_logit_mixing}

\subsection{Ablation of Generation Configuration}
We further analyze DARD with varying generation lengths and block lengths. 
The results in Table~\ref{tab:gen_block_length} show that DARD remains stable across different configurations, indicating that the method is robust to these generation configurations.

\subsection{Qualitative Examples}
We provide qualitative examples comparing DARD with other methods.
As shown in Figure~\ref{fig:qualitative_examples}, WINO, which allows bidirectional attention among all tokens during verification, revises an initially correct answer into an incorrect one by following an erroneous reasoning path in later steps.
In the case of Saber, early incorrect tokens are not properly corrected, and the accumulated errors lead to unreliable subsequent token generation.
In contrast, DARD performs verification in confidence order, leading to a more reliable refinement of the reasoning path toward the correct answer.

\input{tables/gen_block_length}

%% file: figures/5_qualitative.tex
\begin{figure*}[t!]
  \centering
  \includegraphics[width=\textwidth]{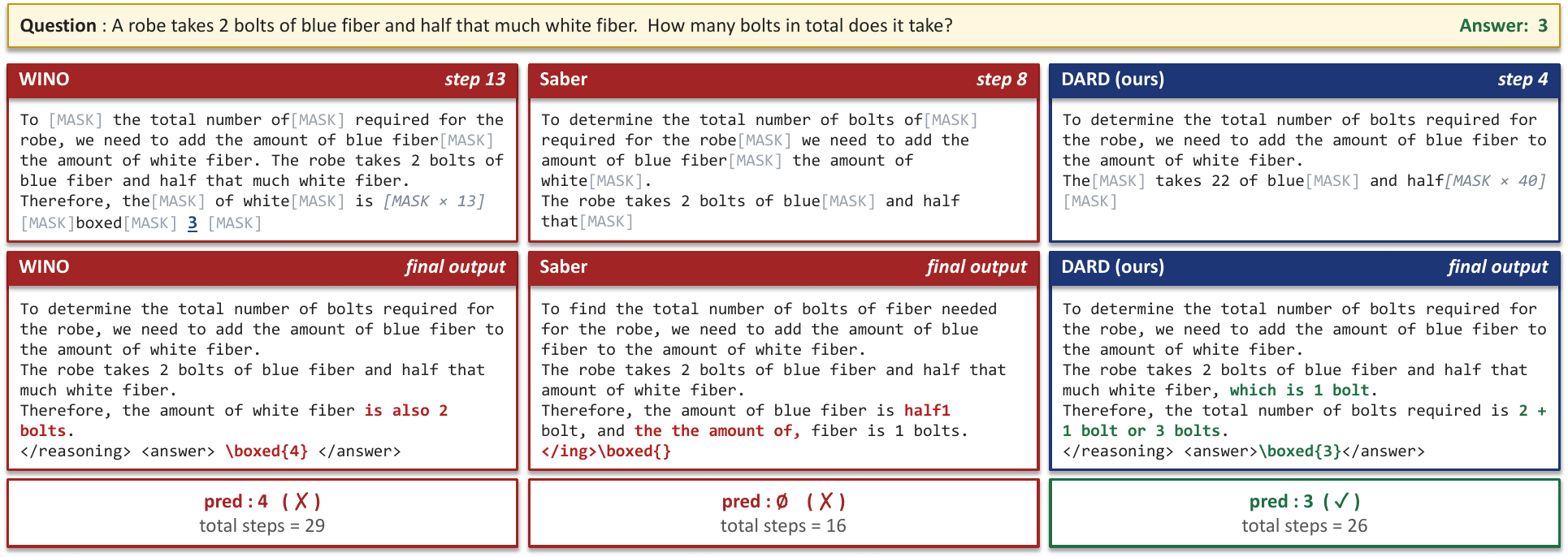}% or \columnwidth
  \caption{
Qualitative comparison of DARD with conventional revocable decoding methods on GSM8K.
}
  \label{fig:qualitative_examples}
\end{figure*}

%% file: tables/attention_mask_ablation.tex
\begin{table}[t]
\centering
\small
\setlength{\tabcolsep}{4pt}
\begin{tabular}{l c c c c}
\toprule
Method & Acc. (\%) & $\Delta$ Acc. (\%) & Steps & $\Delta$ Steps \\
\midrule
$\text{Saber}$               & 33.2 & --  & 81.0 & -- \\
$\text{DARD}_\text{bidir}$   & 31.4 & $-1.8$ & \textbf{58.5} & $-\textbf{22.5}$ \\
$\text{DARD}_\text{l2r}$     & 32.0 & $-1.2$ & 59.3 & $-21.7$ \\
$\text{DARD}_\text{entropy}$ & 33.2 & $+0.0$  & 59.1 & $-21.9$ \\
$\text{DARD}_\text{margin}$  & 34.2 & $+1.0$ & 59.3 & $-21.7$ \\
\midrule
$\text{DARD}$ (ours)         & \textbf{34.6} & $+\textbf{1.4}$ & 59.3 & $-21.7$ \\
\bottomrule
\end{tabular}
\caption{Analysis of attention mask design on DARD. $\Delta$ Acc. and $\Delta$ Steps are computed relative to Saber.}
\label{tab:attention_mask_ablation}
\end{table}

%% file: tables/adaptive_logit_mixing.tex
\begin{table}[t]
\centering
\small
\setlength{\tabcolsep}{3pt} % default is usually 6pt
\begin{tabular}{l c c c c}
\toprule
Method & Acc. (\%) & $\Delta$ Acc. (\%) & Steps & $\Delta$ Steps\\
\midrule
$\text{Saber}$          & 33.2 & --  & 81.0 & -- \\
$\text{DARD}~({w=0.0})$ & 29.2 & $-4.0$ & 82.0 & $+1.0$ \\
$\text{DARD}~({w=0.5})$ & 31.1 & $-2.1$ & 63.6 & $-17.4$ \\
$\text{DARD}~({w=1.0})$ & 33.4 & $+0.2$ & \textbf{57.1} & $-\textbf{23.9}$ \\
\midrule
$\text{DARD}$ (ours) & \textbf{34.6} & $+\textbf{1.4}$ & 59.3 & $-21.7$ \\
\bottomrule
\end{tabular}
\caption{Analysis of logit mixing strategy in DARD. $\Delta$ Acc. and $\Delta$ Steps are computed relative to Saber.}
\label{tab:adaptive_logit_mixing}
\end{table}

%% file: tables/gen_block_length.tex
\begin{table}[t]
\centering
\small
\setlength{\tabcolsep}{5pt}
\begin{tabular}{c c c c}
\toprule
Gen. Len. & Block Len. & Acc. (\%) & Steps \\
\midrule
128 & 128 & 22.0 & 38.9 \\
128 & 64  & 27.0 & 37.1 \\
\midrule
256 & 256 & 22.4 & 62.2 \\
256 & \qquad ~  128 (ours)  & \textbf{34.6} & 59.3 \\
256 & 64  & 32.6 & 61.4 \\
\bottomrule
\end{tabular}
\caption{Ablation of generation length and block length in DARD.}
\label{tab:gen_block_length}
\end{table}

%% file: sections/6_conclusion.tex
\section{Conclusion}
In this paper, we examine a failure mode in revocable decoding for dLLMs.
We observe that incorrectly decoded tokens can distort the context
used for verification, leading to incorrect remasking decisions.
To overcome this limitation, we introduce DARD, a training-free three-state revocable decoding framework that separates decoded tokens according to their confidence.
By decoupling the decoding and verification processes for different states, DARD enables more robust verification without additional training.
Experiments on 6 textual and 6 multimodal benchmarks across three open-source dLLMs show that DARD consistently improves the Pareto frontier of the speed--quality trade-off over recent revocable decoding baselines.
These results highlight the importance of controlling the verification context
for reliable and efficient parallel decoding in dLLMs.

%% file: appendix/appendix.tex
\input{appendix/figures/appendix_experiment_llada15}

\section{Experimental Details}
\label{app:experimental_details}

\subsection{Datasets and Baselines}
We evaluate DARD across a diverse suite of language and vision-language benchmarks.
We evaluate DARD on six language benchmarks--GSM8K~\cite{cobbe2021training_gsm8k}, MATH500~\cite{lightman2023let_math500, hendrycks2021measuring_math_orig}, MBPP~\cite{austin2021program_mbpp}, Countdown~\cite{gandhi2024stream_countdown}, Sudoku~\cite{ye2025beyond_sudoku}, and ARC-Challenge (ARC-C)~\cite{clark2018think_arc_challenge_easy}--covering mathematical, code, arithmetic, logical, and commonsense/science reasoning.
We further evaluate on six vision-language benchmarks--Flickr30K~\cite{young2014flickr30k}, AI2D~\cite{kembhavi2016ai2d}, MATH-Vision~\cite{wang2024mathvision}, MathVista~\cite{lu2024mathvista}, MMMU~\cite{yue2024mmmu}, and ScienceQA~\cite{lu2022scienceqa}--covering image captioning, diagram understanding, visual mathematical reasoning, multi-discipline reasoning, and science question answering.
We compare DARD against standard dLLM decoding, WINO~\cite{hong2025wino}, and Saber~\cite{dong2025saber}, and evaluate them using the validation split of MMMU, the official testmini splits of MathVista and MATH-Vision, the IMG split of ScienceQA, and the 500-sample lite test split of Flickr30K.

\subsection{Implementation Details}
We adopt block decoding~\cite{nie2026llada} for all experiments.
For language benchmarks, we use two open-source dLLMs, LLaDA-8B-Instruct~\cite{nie2026llada} and LLaDA-1.5~\cite{zhu2025llada1.5}; for vision-language benchmarks, we use MMaDA-8B-MixCoT~\cite{yang2026mmada}.
Unless otherwise specified, we use a generation length of $256$ and a block length of $128$ across all methods and benchmarks.
To analyze the speed-quality trade-off, we vary the hyperparameters that control the number of decoding steps.
For DARD, we evaluate all valid pairs of $(\tau_{\mathrm{c}}, \tau_{\mathrm{u}})$ satisfying $\tau_{\mathrm{c}} \leq \tau_{\mathrm{u}}$, using $\tau_{\mathrm{c}} \in \{0.4,0.5,0.6\}$ and $\tau_{\mathrm{u}} \in \{0.6,0.7,0.8\}$ for language tasks, and $\tau_{\mathrm{c}} \in \{0.3,0.4,0.5\}$ and $\tau_{\mathrm{u}} \in \{0.7,0.8,0.9\}$ for vision-language tasks.
We fix the decay factor $\lambda = 0.917$ (i.e., $0.5^{1/8}$, halving the weight every 8 positions) and $p_0 = 0.1$.
For WINO, we fix $\tau_2=0.9$ and evaluate $\tau_1 \in \{0.3, 0.4, 0.5, 0.6, 0.7\}$.
For Saber, we fix $\mu=2$ and vary the minimum number of tokens committed per step as $n \in \{4,5,6,7,8\}$.
All experiments are conducted on NVIDIA RTX A6000 GPUs.
For consistency, we set the sampling temperature to zero and report single-run results.

\section{Additional Experimental Results of DARD}
\subsection{Evaluation on LLaDA 1.5}
\label{app:llada15_results}

To evaluate the generalizability of DARD across models, we further conduct experiments on LLaDA 1.5~\cite{zhu2025llada1.5}.
As shown in Figure~\ref{fig:experiment_llada1_15}, DARD consistently improves the speed-quality trade-off, demonstrating that its effectiveness is not limited to the model used in the main experiments.

\input{appendix/figures/gen_length_ablation}

\subsection{Generalization to longer generations and complex reasoning}
To examine whether DARD remains effective for longer and more complex reasoning trajectories, we evaluate it on MATH-500 with generation budgets of 512 and 1024 tokens (see Figure~\ref{fig:gen_length_ablation}).
The numbers next to the blue-outlined markers indicate the number of meaningful generated tokens after excluding trailing EOT tokens.
DARD consistently achieves a better speed-quality trade-off than competing methods across both generation budgets, demonstrating its effectiveness in mitigating decoding errors caused by long-range dependency failures.

\input{appendix/figures/fine_grained_multimodal_benchmark}

\subsection{Evaluation on fine-grained multimodal benchmarks}

We further evaluate DARD on the fine-grained multimodal benchmarks MMVP~\cite{tong2024MMVP}, BLINK~\cite{fu2024blink}, and HRBench~\cite{wang2025HRBench}, using the same experimental configuration as in the main experiments (see Figure~\ref{fig:fine_grained_multimodal_benchmark}).
Across a range of confidence thresholds $\tau_c$ and $\tau_u$, DARD consistently achieves a better speed--quality Pareto frontier than the baseline methods.
These results demonstrate that DARD generalizes across diverse multimodal benchmarks and remains robust to threshold selection.

\section{Additional Analysis of DARD}
\subsection{Analysis of Adaptive Logit Mixing}

\paragraph{Qualitative illustration of adaptive logit mixing.}
To illustrate the effect of the adaptive mixing weight on the decoding outcome, Table~\ref{tab:candidate_transition_example} shows an intermediate decoding stage for the caption \textit{``A man in a black shirt is reading a book''}.
At step~13, position~7 is decoded as \textit{``on''} and assigned to the $\mathcal{C}$ state.
In the following step, conditioning on this candidate token leads the original path to predict \textit{``sits''} at position~6, yielding the plausible phrase \textit{``sits on''}.
By contrast, the shadow path excludes this candidate token from its context and predicts \textit{``is''} at position~6.
After \textit{``on''} at position~7 is demoted to the masked state, the mixing weight for position~6 is set to a low value of $w=0.24$. 
Consequently, assigning greater weight to the shadow path leads the mixed prediction to select \textit{``is''}, thereby avoiding the erroneous prediction \textit{``sits''} induced by the incorrect candidate context.

\newcommand{\cand}[1]{% 
    \ensuremath{\langle\text{#1}\rangle}% 
} 
\newcommand{\demote}{%
    \textbf{\texttt{[M]}}\,\ensuremath{\boldsymbol{\downarrow}}%
}
\begin{table*}[t]
    \centering
    \small
    \setlength{\tabcolsep}{3.5pt}
    \renewcommand{\arraystretch}{1.15}

    \begin{tabular}{l*{11}{c}}
        \toprule
        Position
        & 0 & 1 & 2 & 3 & 4 & 5 & 6 & 7 & 8 & 9 & 10 \\
        \midrule

        Step 13
        & A
        & man
        & \cand{wearing}
        & a
        & \cand{black}
        & shirt
        & \texttt{[M]}
        & \cand{on}
        & \cand{a}
        & book
        & . \\

        Step 14
        & A
        & man
        & \demote
        & a
        & \cand{black}
        & shirt
        & \textbf{is}
        & \demote
        & \cand{a}
        & book
        & . \\

        \midrule

        Final
        & A
        & man
        & in
        & a
        & black
        & shirt
        & is
        & reading
        & a
        & book
        & . \\

        \bottomrule
    \end{tabular}

        \caption{
        Illustration of how the adaptive mixing weight operates during MMaDA generation on Flickr30K. 
        $\langle\cdot\rangle$ denotes a token in the candidate state, and $\downarrow$ indicates demotion to the masked state.
    }
    \label{tab:candidate_transition_example}
\end{table*}

\paragraph{Statistics of mixing weights $w_t^i$ during decoding.}
We additionally report the proportion of observed weights $w_t^i$ during decoding: GSM8K with LLaDA and ScienceQA with MMaDA.
\begin{table*}[t]
    \centering
    \small
    \setlength{\tabcolsep}{5pt}
    \renewcommand{\arraystretch}{1.1}
    \begin{tabular}{lcccc}
        \toprule
        Model / Dataset
        & $w_t^i < 0.5$
        & $0.5 \leq w_t^i < 0.7$
        & $0.7 \leq w_t^i < 0.9$
        & $w_t^i \geq 0.9$ \\
        \midrule
        LLaDA / GSM8K
        & 4.7\%
        & 5.1\%
        & 6.8\%
        & 83.3\% \\

        MMaDA / ScienceQA
        & 10.7\%
        & 5.1\%
        & 4.9\%
        & 79.2\% \\
        \bottomrule
    \end{tabular}
        \caption{
        Empirical distribution of adaptive mixing weights $w_t^i$ observed during
        decoding. Each entry reports the percentage of all observed $w_t^i$ values
        that fall within the corresponding range.
    }
    \label{tab:adaptive_weight_distribution}
\end{table*}
As shown in Table~\ref{tab:adaptive_weight_distribution}, $w_t^i$ exceeds 0.9 in most decoding steps, indicating that the model generally leverages $\mathcal{C}$ tokens as context during token prediction.
For a small fraction of tokens, $w_t^i$ is assigned much lower values ($<0.5$), reducing the influence of $\mathcal{C}$ token context when it is not reliable.
These results illustrate why using a single fixed weight is not optimal. 
Although $\mathcal{C}$ tokens are valid and useful as context in most cases, assigning them a uniformly high weight can cause the model to continue relying on $\mathcal{C}$ token context even when some $\mathcal{C}$ tokens are demoted, which can lead to incorrect decoding decisions.

\begin{table*}[t]
    \centering
    \small
    \setlength{\tabcolsep}{5pt}
    \renewcommand{\arraystretch}{1.1}
    \begin{tabular}{lllccccc}
        \toprule
        Model / Dataset
        & Metric
        & Ordering
        & $K{=}1$
        & $K{=}2$
        & $K{=}4$
        & $K{=}8$
        & $K{=}16$ \\
        \midrule

        \multirow{4}{*}{\makecell[l]{LLaDA / GSM8K}}
        & \multirow{2}{*}{$\mathrm{Overlap@}K$ ($\uparrow$)}
        & $\pi^{conf}$
        & \textbf{1.00}
        & \textbf{0.86}
        & \textbf{0.82}
        & \textbf{0.81}
        & \textbf{0.81} \\
        &
        & $\pi^{l2r}$
        & 0.24
        & 0.36
        & 0.49
        & 0.60
        & 0.67 \\

        \cmidrule(lr){2-8}

        &
        \multirow{2}{*}{$\mathrm{RankDiff@}K$ ($\downarrow$)}
        & $\pi^{conf}$
        & \textbf{0.00}
        & \textbf{0.74}
        & \textbf{1.74}
        & \textbf{3.27}
        & \textbf{5.62} \\
        &
        & $\pi^{l2r}$
        & 15.38
        & 13.57
        & 12.52
        & 12.40
        & 13.03 \\

        \midrule

        \multirow{4}{*}{\makecell[l]{MMaDA / Flickr30K}}
        & \multirow{2}{*}{$\mathrm{Overlap@}K$ ($\uparrow$)}
        & $\pi^{conf}$
        & \textbf{1.00}
        & \textbf{0.73}
        & \textbf{0.64}
        & \textbf{0.64}
        & \textbf{0.69} \\
        &
        & $\pi^{l2r}$
        & 0.04
        & 0.05
        & 0.08
        & 0.11
        & 0.18 \\

        \cmidrule(lr){2-8}

        &
        \multirow{2}{*}{$\mathrm{RankDiff@}K$ ($\downarrow$)}
        & $\pi^{conf}$
        & \textbf{0.00}
        & \textbf{3.08}
        & \textbf{5.68}
        & \textbf{8.06}
        & \textbf{10.06} \\
        &
        & $\pi^{l2r}$
        & 46.15
        & 47.20
        & 46.83
        & 46.83
        & 43.84 \\

        \bottomrule
    \end{tabular}
        \caption{
        Comparison of confidence-based and left-to-right orderings with the one-token-per-step decoding trajectory.
        Higher $\mathrm{Overlap@}K$ and lower $\mathrm{RankDiff@}K$ indicate
        better alignment. 
        All metrics are averaged over samples and decoding steps.
    }
    \label{tab:ordering_alignment}
\end{table*}

\subsection{Comparing Confidence-Based Ordering with Subsequent Decoding Trajectories}
\label{app:conf_order}

To empirically examine the intuition that confidence-based ordering reflects the subsequent decoding trajectory, we conduct an analysis using LLaDA on GSM8K and MMaDA on Flickr30K.
At each decoding step, we compare two candidate orderings of the remaining masked positions against a reference ordering.
We define the confidence-based ordering $\pi^{\mathrm{conf}}$ by ranking the remaining masked positions in descending order of their confidence scores, and the left-to-right ordering $\pi^{\mathrm{l2r}}$ by ranking them from left to right.
We obtain the reference ordering $\pi^{\mathrm{ref}}$ by decoding one token per step until all remaining masked positions are decoded and recording the resulting order.

We compare each candidate ordering with the reference ordering using two metrics, $\mathrm{Overlap@}K$ and $\mathrm{RankDiff@}K$.
For $\pi \in \{\pi^{\mathrm{conf}}, \pi^{\mathrm{l2r}}\}$, the metrics are defined as follows:
$$\mathrm{Overlap@}K \!=\!\frac{\left|\operatorname{TopK}(\pi) \cap \operatorname{TopK}(\pi^{\mathrm{ref}})\right|}{K},$$

\[ \mathrm{RankDiff@}K = \frac{1}{K}\sum\limits_{\scriptstyle i \in \operatorname{TopK}(\pi)}\left| \pi(i) - \pi^{\mathrm{ref}}(i) \right|. \]

\noindent 
Here, $\operatorname{TopK}(\pi)$ denotes the set of the first $K$ positions under ordering $ \pi$, and $\pi(i)$ denotes the rank of position $i$ in that ordering. 
$\mathrm{Overlap@}K$ measures the fraction of positions shared by the top-$K$ sets of $\pi$ and $\pi^{\mathrm{ref}}$.
$\mathrm{RankDiff@}K$ measures the average absolute difference between the ranks of these positions in $\pi$ and $\pi^{\mathrm{ref}}$.

As shown in Table~\ref{tab:ordering_alignment}, $\pi^{\mathrm{conf}}$ aligns substantially better with the reference ordering $\pi^{\mathrm{ref}}$ than $\pi^{\mathrm{l2r}}$ does across both models.
Since DARD decodes an average of 6.0 and 8.8 tokens per step with LLaDA and MMaDA, respectively, the high $\mathrm{Overlap@}8$ values further indicate that our confidence-based ordering closely reflects the subsequent decoding order.

\subsection{Accuracy Gains Beyond Efficiency}
The primary goal of DARD is to improve the speed–quality Pareto frontier of dLLM decoding. 
At the same time, DARD can also improve peak accuracy by iteratively verifying and refining generated tokens using richer context formed by the model’s own predictions, instead of permanently fixing tokens predicted under limited context. 
This resembles self-reflection~\cite{madaan2023self} in AR-LLMs, where the model iteratively revisits and refines its own generated output before finalizing the answer.
To make verification more robust, DARD selectively uses only reliable decoded tokens as context, which is consistent with prior work~\cite{huang2024large} showing that self-reflection with unreliable context can even degrade performance.

To empirically demonstrate that DARD improves absolute task performance, we additionally compare it with the best-performing configuration of the default decoding setting (i.e., one-token-per-step decoding) on six datasets. 
As shown in Tables~\ref{tab:accuracy_steps_llada} and~\ref{tab:accuracy_steps_mmada}, DARD achieves higher peak accuracy than this configuration across all six datasets.

\begin{table}[t]
    \centering
    \small
    \setlength{\tabcolsep}{3pt}
    \renewcommand{\arraystretch}{1.05}
    \begin{tabular}{llcccc}
        \toprule
        Benchmark
        & Method
        & \makecell{Acc.\\(\%)}
        & Steps
        & \makecell{Acc.\\gain (\%)}
        & \makecell{Step\\reduction} \\
        \midrule

        \multirow{2}{*}{ARC-C}
        & Default & 52.17 & 256 & -- & -- \\
        & DARD & \textbf{81.61} & \textbf{35} & \textit{+56.4} & \textit{7.2$\times$} \\

        \addlinespace[2pt]

        \multirow{2}{*}{Countdown}
        & Default & 29.00 & 256 & -- & -- \\
        & DARD & \textbf{34.51} & \textbf{85} & \textit{+19.0} & \textit{3.0$\times$} \\

        \addlinespace[2pt]

        \multirow{2}{*}{GSM8K}
        & Default & 73.77 & 256 & -- & -- \\
        & DARD & \textbf{77.86} & \textbf{48} & \textit{+5.5} & \textit{5.3$\times$} \\

        \bottomrule
    \end{tabular}
        \caption{
        Comparison of peak-accuracy configurations for the default setting and DARD using LLaDA.
    }
    \label{tab:accuracy_steps_llada}
\end{table}

\begin{table}[t]
    \centering

    \small
    \setlength{\tabcolsep}{3pt}
    \renewcommand{\arraystretch}{1.05}
    \begin{tabular}{llcccc}
        \toprule
        Benchmark
        & Method
        & \makecell{Acc.\\(\%)}
        & Steps
        & \makecell{Acc.\\gain (\%)}
        & \makecell{Step\\reduction} \\
        \midrule

        \multirow{2}{*}{MMMU-val}
        & Default & 20.56 & 256 & -- & -- \\
        & DARD & \textbf{22.89} & \textbf{34} & \textit{+11.3} & \textit{7.5$\times$} \\

        \addlinespace[2pt]

        \multirow{2}{*}{AI2D}
        & Default & 55.25 & 256 & -- & -- \\
        & DARD & \textbf{56.15} & \textbf{37} & \textit{+1.6} & \textit{7.0$\times$} \\

        \addlinespace[2pt]

        \multirow{2}{*}{ScienceQA}
        & Default & 41.89 & 256 & -- & -- \\
        & DARD & \textbf{44.67} & \textbf{25} & \textit{+6.6 }& \textit{10.3$\times$} \\

        \bottomrule
    \end{tabular}
        \caption{
        Comparison of peak-accuracy configurations for the default setting and DARD using MMaDA.
    }
    \label{tab:accuracy_steps_mmada}
\end{table}

\begin{table*}[t]
    \centering

    \small
    \setlength{\tabcolsep}{4pt}
    \renewcommand{\arraystretch}{1.1}

    \begin{tabular}{cccccc}
        \toprule
        \multirow{2}{*}{$\tau_c$}
        & \multicolumn{5}{c}{$\tau_u$} \\
        \cmidrule(lr){2-6}
        & 0.5 & 0.6 & 0.7 & 0.8 & 0.9 \\
        \midrule

        0.2
        & 25.2 {\color{gray}(32.81)}
        & 26.0 {\color{gray}(36.76)}
        & 24.2 {\color{gray}(39.84)}
        & 23.8 {\color{gray}(43.76)}
        & 29.0 {\color{gray}(50.11)} \\

        0.3
        & 25.8 {\color{gray}(32.37)}
        & 25.4 {\color{gray}(34.54)}
        & 27.2 {\color{gray}(37.61)}
        & 26.2 {\color{gray}(40.90)}
        & 28.6 {\color{gray}(44.82)} \\

        0.4
        & 30.0 {\color{gray}(34.45)}
        & 27.6 {\color{gray}(37.93)}
        & 28.0 {\color{gray}(41.07)}
        & 28.2 {\color{gray}(44.91)}
        & 30.8 {\color{gray}(51.75)} \\

        0.5
        & --
        & 30.0 {\color{gray}(39.54)}
        & 28.2 {\color{gray}(45.50)}
        & 31.0 {\color{gray}(51.85)}
        & 32.0 {\color{gray}(59.46)} \\

        0.6
        & --
        & --
        & 31.4 {\color{gray}(48.72)}
        & 34.4 {\color{gray}(56.94)}
        & 34.0 {\color{gray}(66.76)} \\

        0.7
        & --
        & --
        & --
        & 32.8 {\color{gray}(61.42)}
        & 32.2 {\color{gray}(74.32)} \\

        \bottomrule
    \end{tabular}
        \caption{
        Effects of the threshold hyperparameters $\tau_c$ and $\tau_u$ on the speed--quality trade-off of LLaDA on MATH-500.
        Each cell reports accuracy (\%), with the average number of decoding steps shown in gray parentheses.
    }
    \label{tab:tau_ablation}
\end{table*}

\subsection{Effects of Thresholds on the Speed--Quality Trade-off}

\paragraph{Robustness across models and datasets.}
The thresholds $\tau_c$ and $\tau_u$ adjust the speed--quality trade-off in DARD by controlling how conservatively tokens are selected and verified.
As shown in Figures~\ref{fig:experiment_llm}, ~\ref{fig:experiment_lvlm}, and~\ref{fig:experiment_llada1_15}, varying these thresholds produces different speed--quality operating points, with most DARD configurations outperforming the Pareto frontier established by prior baselines.
Specifically, the fixed threshold settings $(\tau_c,\tau_u) \in {(0.5, 0.8), (0.6, 0.7)}$ for LLaDA and $(\tau_c,\tau_u) \in {(0.3, 0.9), (0.4, 0.9)}$ for MMaDA remain above the Pareto frontier of prior methods across all five datasets evaluated for each architecture.
These results suggest that DARD is robust to threshold choices and does not require task-specific tuning.

\paragraph{Roles of $\tau_c$ and $\tau_u$ in decoding.}
To further analyze how $\tau_c$ and $\tau_u$ affect the decoding process, we report results for different threshold settings on MATH-500 using LLaDA in Table~\ref{tab:tau_ablation}. 
The results show that increasing $\tau_u$ generally increases the number of decoding steps while improving task performance.
This is because a larger $\tau_u$ requires higher confidence before a token can be used as a reliable context, making the decoding process more conservative and reducing error propagation.
In addition, decreasing $\tau_c$ reduces the number of decoding steps but often leads to lower accuracy. 
This is because a smaller $\tau_c$ relaxes the criterion for assigning tokens to the candidate state, allowing more low-confidence tokens to be used during verification.
These noisy candidate tokens can provide unreliable context and hurt performance.

\paragraph{Practical guidance for threshold selection.}
Consistent with our design intent, the above observations provide practical guidance for selecting the thresholds.
For accuracy-oriented settings, both thresholds should be moderately high so that only sufficiently confident tokens are used in the verification process, such as $(\tau_c,\tau_u)=(0.6,0.8)$ or $(0.6,0.9)$. 
For speed-oriented settings, setting $\tau_u$ to a lower value and choosing $\tau_c$ close to it, such as $(\tau_c,\tau_u)=(0.4,0.5)$ or $(0.5,0.6)$, can reduce the number of decoding steps while maintaining competitive performance.

\subsection{Analysis of Computational Efficiency and Overhead}

\paragraph{End-to-end TPS and task performance.}
In Figures~\ref{fig:experiment_llm},~\ref{fig:experiment_lvlm}, and ~\ref{fig:experiment_llada1_15}, we report the tokens-per-second (TPS) values next to the red outlines. 
TPS is defined as the number of generated output tokens divided by the total wall-clock decoding time. 
Since the generated output tokens are fixed in our comparison, a higher TPS directly corresponds to shorter end-to-end wall-clock time. 

To make the comparison more explicit, we summarize representative results in Tables~\ref{tab:tps_performance_llada} and~\ref{tab:tps_performance_mmada}, reporting both TPS and the corresponding task performance. 
In most cases, DARD achieves higher TPS and better task performance simultaneously. 
For benchmarks where DARD does not achieve the highest TPS, it provides a clear gain in task performance; for example, Saber is faster on ARC-C and Countdown, but underperforms DARD by 7.4 and 5.6 percentage points, respectively.

\begin{table}[t]
    \centering

    %------------------------------------------------
    % LLaDA
    %------------------------------------------------
    \resizebox{\linewidth}{!}{
    \begin{tabular}{c cc cc cc}
        \toprule
        \multirow{2}{*}{Method}
        & \multicolumn{2}{c}{ARC-C}
        & \multicolumn{2}{c}{GSM8K}
        & \multicolumn{2}{c}{Countdown} \\
        \cmidrule(lr){2-3}
        \cmidrule(lr){4-5}
        \cmidrule(lr){6-7}
        & TPS & Acc.
        & TPS & Acc.
        & TPS & Acc. \\
        \midrule
        Saber
        & \textbf{69.0} & 72.2
        & 30.5 & 74.7
        & \textbf{24.7} & 28.9 \\

        WINO
        & 25.4 & 78.9
        & 30.6 & \textbf{77.8}
        & 12.1 & 31.8 \\

        DARD
        & 36.4 & \textbf{79.6}
        & \textbf{34.5} & 77.2
        & 15.5 & \textbf{34.5} \\
        \bottomrule
    \end{tabular}
    }
        \caption{
        Comparison of inference throughput and task performance for LLaDA.
        Higher values are better for all metrics.
    }
    \label{tab:tps_performance_llada}
\end{table}

\begin{table}[t]
    \centering

    %------------------------------------------------
    % MMaDA
    %------------------------------------------------
    \resizebox{\linewidth}{!}{
    \begin{tabular}{c cc cc cc}
            \toprule
            \multirow{2}{*}{Method}
            & \multicolumn{2}{c}{AI2D}
            & \multicolumn{2}{c}{ScienceQA}
            & \multicolumn{2}{c}{Flickr30K} \\
            \cmidrule(lr){2-3}
            \cmidrule(lr){4-5}
            \cmidrule(lr){6-7}
            & TPS & Acc.
            & TPS & Acc.
            & TPS & CIDEr \\
            \midrule
            Saber
            & 11.3 & 53.3
            & 13.5 & \textbf{44.7}
            & 12.2 & 54.8 \\

            WINO
            & 11.4 & 53.9
            & 14.0 & 43.5
            & \textbf{18.5} & 57.1 \\

            DARD
            & \textbf{14.5} & \textbf{55.5}
            & \textbf{16.0} & \textbf{44.7}
            & 18.2 & \textbf{59.2} \\
            \bottomrule
        \end{tabular}
    }
        \caption{
        Comparison of inference throughput and task performance for MMaDA.
        Higher values are better for all metrics.
    }
    \label{tab:tps_performance_mmada}
\end{table}

\paragraph{Single-step inference latency.}
To better quantify the computational overhead, we further report the wall-clock time of a single forward inference pass. 
As shown in Table~\ref{tab:latency_memory}, DARD introduces negligible overhead compared with WINO for a single forward inference pass. 
Although this overhead is slightly larger than Saber’s, DARD requires substantially fewer decoding steps, which results in shorter total wall-clock decoding time.

\paragraph{Peak GPU memory overhead.}
We also report the peak GPU memory overhead of DARD. 
The additional memory cost mainly comes from the shadow block sequence, whose size depends on the block length. 
Therefore, we report results under two settings with the same generation length of 256: the default setting used in the manuscript with block length 128, and the most memory-intensive setting with block length 256. 
The results show that the increase in peak GPU memory is less than 6.3\% in the worst case, indicating that DARD incurs only negligible peak memory overhead. 
This is because the added shadow block accounts for only a small fraction of the total sequence, which includes image tokens, system/text prompt tokens, and generated tokens. 
In addition, peak GPU memory is largely dominated by the model parameters, making the extra memory from the shadow block relatively small.

In summary, these results show that DARD improves the speed–quality trade-off by achieving strong task performance with competitive or shorter end-to-end wall-clock time, while incurring negligible peak GPU memory overhead.

\begin{table}[t]
    \centering

    \small
    \setlength{\tabcolsep}{4pt}
    \begin{tabular}{lccc}
        \toprule
        Method
        & \makecell[c]{Block\\length}
        & \makecell[c]{1-step latency\\(ms)}
        & \makecell[c]{GPU memory \\(GB)} \\
        \midrule
        Default & 256  & 137.7 & 17.21 \\
        \midrule
        Saber   & 256  & 141.1 & 17.62 \\
        WINO    & 128 & 183.9 & 17.45 \\
        WINO    & 256 & 229.6 & 17.96 \\
        \midrule
        DARD    & 128 & 186.8 & 17.65 \\
        DARD    & 256 & 232.0 & 18.29 \\
        \bottomrule
    \end{tabular}
        \caption{
        Single-step inference latency and peak GPU memory usage of LLaDA on MATH-500. All methods use a generation length of 256.
    }
    \label{tab:latency_memory}
\end{table}

\paragraph{Attention overhead of the shadow sequence.}

We additionally analyze the theoretical attention overhead introduced by the shadow sequence.
DARD duplicates only the block currently being decoded instead of duplicating the full input sequence.
The full sequence consists of $S$ system prompt tokens, $V$ visual tokens, $P$ text prompt tokens, and $L$ generation tokens, with a total length of $N=S+V+P+L$.
Under the block decoding framework, the generation tokens are further divided into $K$ decoding blocks of length $B$ (i.e., $L = BK$). 
Therefore, the attention cost with the additional shadow sequence increases from $O(N^2)$ to $O((N+B)^2)$, where $B \ll N$.
Moreover, dLLM decoding can use prefix caching for the system prompt tokens, visual tokens, and text prompt tokens. 
With prefix caching, the attention cost during generation is better characterized as increasing from $O(LN)$ to $O((L+B)(N+B))$.
In our main setting, the block length is 128, whereas a single image already contains around 1024 visual tokens.
Therefore, the shadow sequence introduces only a small relative overhead, which further decreases in high-resolution image or video settings as $B/N$ becomes smaller.

\section{Comparison with Concurrent Work on Parallel Decoding}
\label{comp_with_con_work}
Combination errors and joint inconsistencies, such as \textit{``Los Diego''}, are commonly observed in parallel decoding for diffusion language models. 
Rejection Mixing~\cite{ye2026rejection}, DAPD~\cite{kim2026dapd}, and EB-Sampler~\cite{ben2026EBsampler} also aim to mitigate this problem, but differ from DARD in their treatment of dependencies among simultaneously predicted tokens.
Rejection Mixing addresses this problem by using soft embeddings that encode information from multiple vocabulary tokens predicted at intermediate steps. 
DAPD constructs a dependency graph based on attention maps and selects tokens that are weakly dependent on each other. 
EB-Sampler controls the error upper bound of the joint probability that arises when multiple tokens are selected simultaneously. 
Compared with these methods, DARD explicitly controls the conditioning context of each token prediction so that the probability of multiple tokens can be represented through an ordered conditional factorization.

\section{Modeling Token Dependencies during Verification}
\label{dependency_details}
The mask design of DARD ensures that, when verifying a $\mathcal{C}$ token, the model can attend only to higher-confidence vocabulary tokens and $\mathcal{U}$ tokens.
Thus, for the $k$-th $\mathcal{C}$ token under the ordering $\pi_t$, we have
\begin{equation}
\small
p_\theta\left(s_{t+1}^{\pi_t(k)} \mid \tilde{\mathbf x_t}  ; M\right)\approx p_\theta\left(s_{t+1}^{\pi_t(k)}\mid\hat{x}_t^{\pi_t(1:{k-1})},\hat{x}_t^{\mathcal{U}_t}\right),
\end{equation}
where $\pi_t(k)$ denotes the $k$-th token in the $\mathcal{C}$ state ordering, $k = 1,\dots,|\mathcal{C}_t|$.

When a token is accepted through verification, its verification outcome matches the previously predicted token, i.e., $s_{t+1}^{\pi_t(i)} = \hat{x}_t^{\pi_t(i)}$. Under this condition, the joint probability over the first $k$ $\mathcal{C}$ tokens can be expressed by the chain rule as
\begin{equation}
\small
\begin{aligned}
p_\theta\!\left(
    s_{t+1}^{\pi_t(1:k)}
    \,\middle|\,
    \hat{x}_t^{\mathcal{U}_t}
\right)
&=
p_\theta\!\left(
    s_{t+1}^{\pi_t(1)}
    \,\middle|\,
    \hat{x}_t^{\mathcal{U}_t}
\right)
\\[-1mm]
&\qquad
\prod_{i=2}^{k}
p_\theta\!\left(
    s_{t+1}^{\pi_t(i)}
    \,\middle|\,
    s_{t+1}^{\pi_t(1:i-1)},
    \hat{x}_t^{\mathcal{U}_t}
\right).
\end{aligned}
\end{equation}

By incorporating token dependencies directly into the verification process, DARD allows the verification outcome to reflect the joint distribution over committed tokens.

\section{Handling Contamination Propagated to Low-Confidence Tokens}
In revocable decoding frameworks, erroneous tokens cannot be completely excluded from the conditioning context because their correctness cannot be determined before verification.
Nevertheless, DARD effectively reduces the propagation of erroneous context by enforcing information flow along the confidence order, i.e., from high-confidence tokens to low-confidence tokens.
As a result, even if an erroneous token contaminates lower-confidence tokens, higher-confidence tokens remain unaffected and can provide reliable context in subsequent decoding steps.

Although a high-confidence token may still be erroneous and affect lower-confidence tokens, DARD mitigates the resulting error propagation through its iterative verification process.
Once an erroneous token is identified and demoted, it is excluded from the conditioning context in subsequent verification steps.
As a result, the confidence of tokens affected by the demoted token decreases, making them more likely to be demoted as well.

This behavior is illustrated in Figure~\ref{fig:appendix_qualitative_5}.
At step 7, the incorrect tokens \textit{``football''} and \textit{``field''} both enter the $\mathcal{C}$ state, with a higher confidence score for \textit{``field''}.
At step 8, \textit{``football''} uses \textit{``field''} as conditioning context and promotes to $\mathcal{U}$.
In contrast, \textit{``field''} cannot use \textit{``football''} as context because \textit{``football''} has lower confidence.
Instead, it relies on context from other tokens, such as \textit{``talking on''}, and thereby demotes to $\mathcal{M}$.
Once \textit{``field''} is removed from the conditioning context, \textit{``football''} can no longer rely on it. 
Its confidence consequently decreases, causing it to demote at step 9.
Thus, even when an error has propagated to lower-confidence tokens, DARD can progressively correct the resulting errors through subsequent verification steps.

\section{Additional Qualitative Examples}
We provide two types of qualitative comparisons.
\Cref{fig:appendix_qualitative_1,fig:appendix_qualitative_2} compare DARD with other methods at intermediate decoding steps and in the final outputs.
\Cref{fig:appendix_qualitative_3,fig:appendix_qualitative_4} illustrate the differences between DARD and conventional revocable decoding in the correction of erroneous tokens.
These comparisons demonstrate that DARD effectively resolves decoding errors and reduces their propagation during verification.

\section{Pseudocode for DARD}
We present the full DARD procedure, detailing its dependency-aware token verification and revocation process (see Algorithm~\ref{alg:state_transition}).

\section{Ethics Statement}
All open-source datasets and publicly available model weights used in this work are used in accordance with their respective licenses, and intended research or evaluation purposes.
The proposed DARD implementation is intended to support reproducible research on efficient dLLM decoding.

\newpage

\input{appendix/figures/appendix_qualitative_5}

\clearpage

\input{appendix/figures/appendix_qualitative_1}
\input{appendix/figures/appendix_qualitative_2}

\clearpage

\input{appendix/figures/appendix_qualitative_3}
\input{appendix/figures/appendix_qualitative_4}

\input{appendix/tables/algorithm}

%% file: appendix/figures/appendix_experiment_llada15.tex
\twocolumn[{%
  % \centering
  \includegraphics[width=\textwidth]{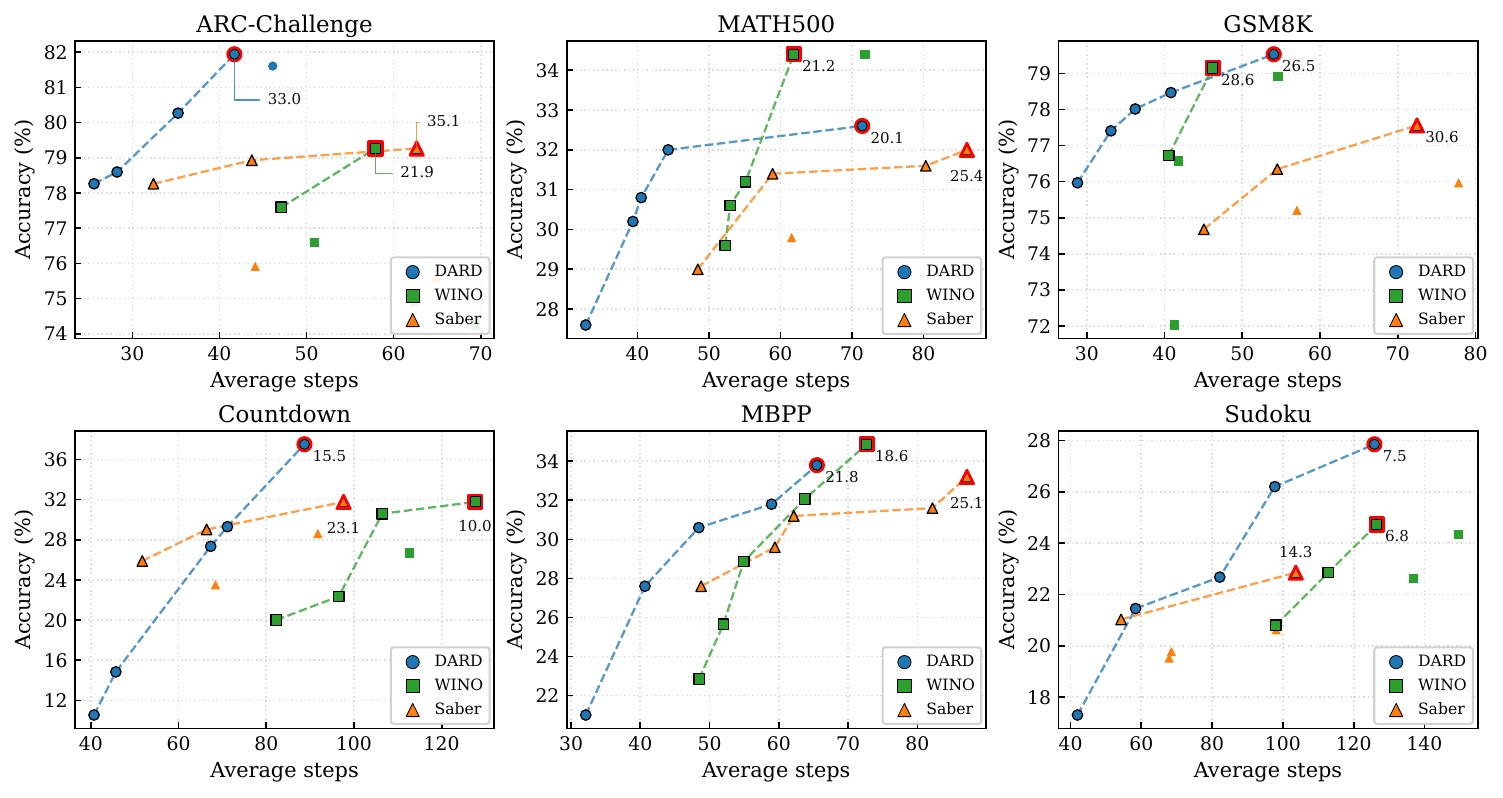}\\[6pt]
  \refstepcounter{figure}%
  Figure~\thefigure:~LLaDA-1.5 results on six language benchmarks. Each plot compares task performance against the average number of decoding steps for each method. For DARD, we show the five configurations closest to each benchmark's accuracy-step Pareto frontier. Dashed curves denote the Pareto frontier of each method. For each method, the highest-accuracy configuration is highlighted with a red outline, with its TPS annotated alongside the point.
  \label{fig:experiment_llada1_15}
  \vspace{1em}
}]

%% file: appendix/figures/gen_length_ablation.tex
\begin{figure}[t!]
  \centering
    \includegraphics[width=\linewidth]{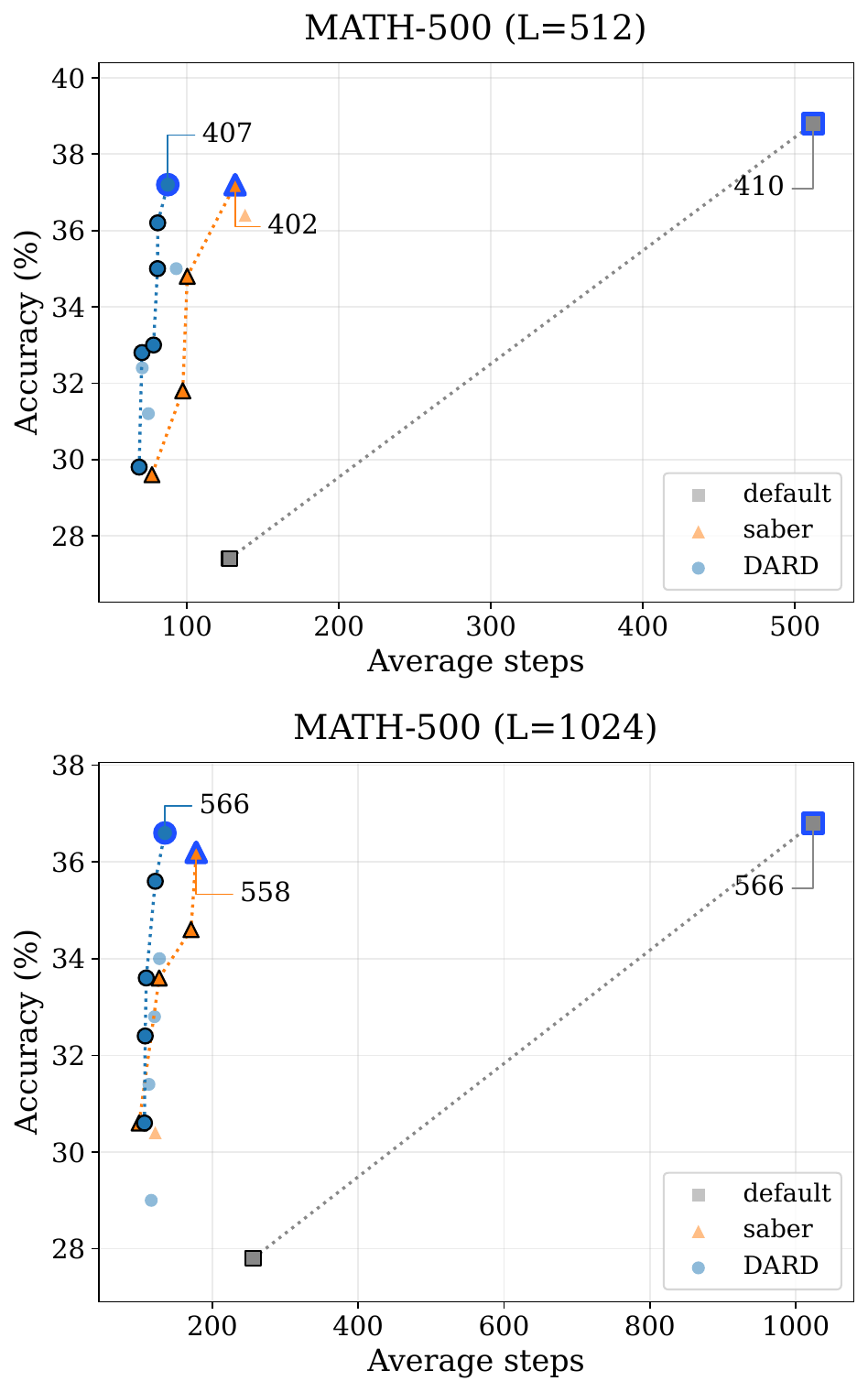}
  \caption{Evaluation with different generation budgets on MATH-500.
  }
% \vspace{-0.4cm}
  \label{fig:gen_length_ablation}
\end{figure}

%% file: appendix/figures/fine_grained_multimodal_benchmark.tex
\begin{figure*}[t!]
  \centering
  \includegraphics[width=\textwidth]{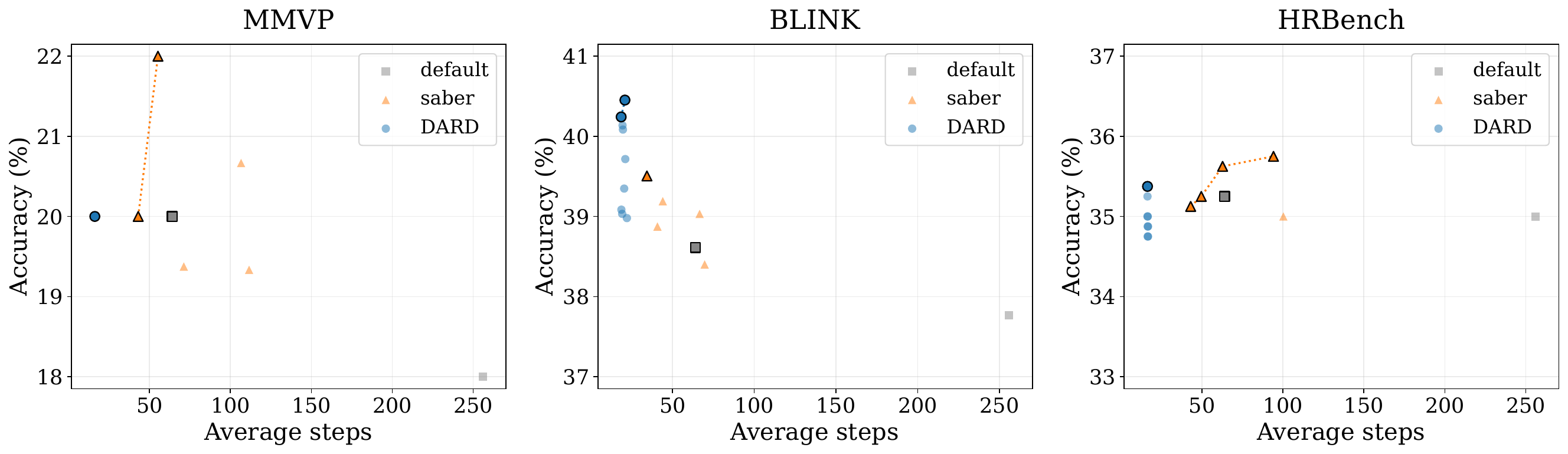}% or \columnwidth
  \caption{Fine-grained multimodal benchmark results.}
  \label{fig:fine_grained_multimodal_benchmark}
\end{figure*}

%% file: appendix/figures/appendix_qualitative_5.tex
\begin{figure*}[t!]
  \centering
  \includegraphics[width=\textwidth]{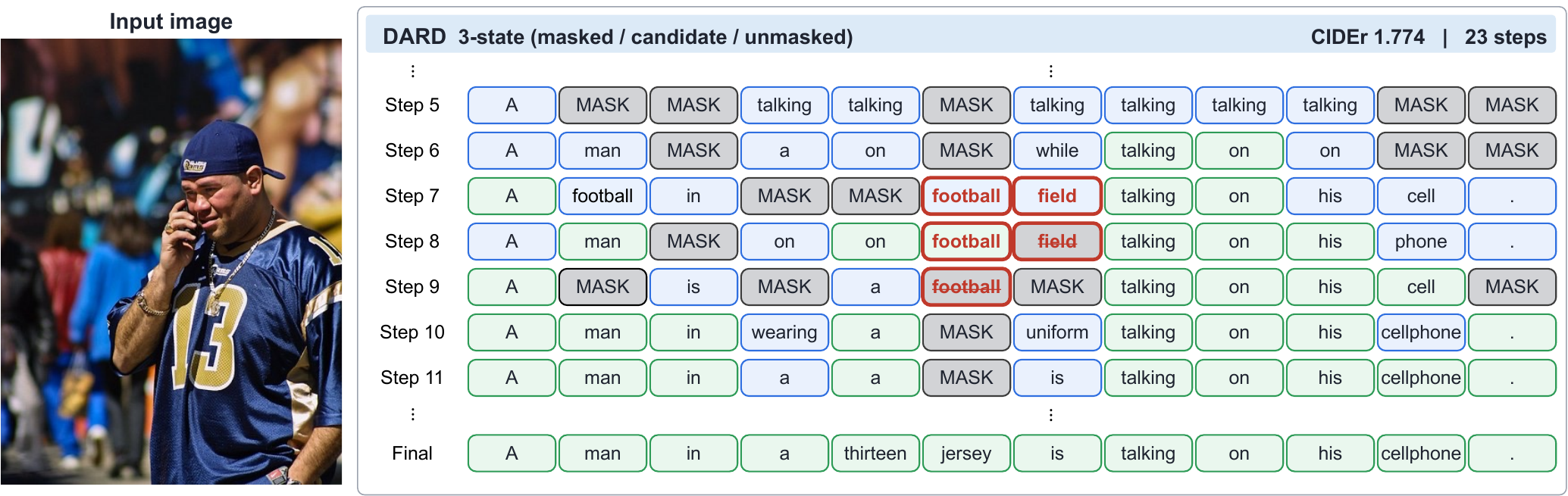}% or \columnwidth
  \caption{
Qualitative example illustrating how DARD effectively mitigates error propagation through iterative verification by demoting erroneous tokens, removing them from subsequent conditioning contexts, and progressively identifying tokens whose confidence depends on those errors. }
  \label{fig:appendix_qualitative_5}
\end{figure*}

%% file: appendix/figures/appendix_qualitative_1.tex
\begin{figure*}[t!]
  \centering
  \includegraphics[width=\textwidth]{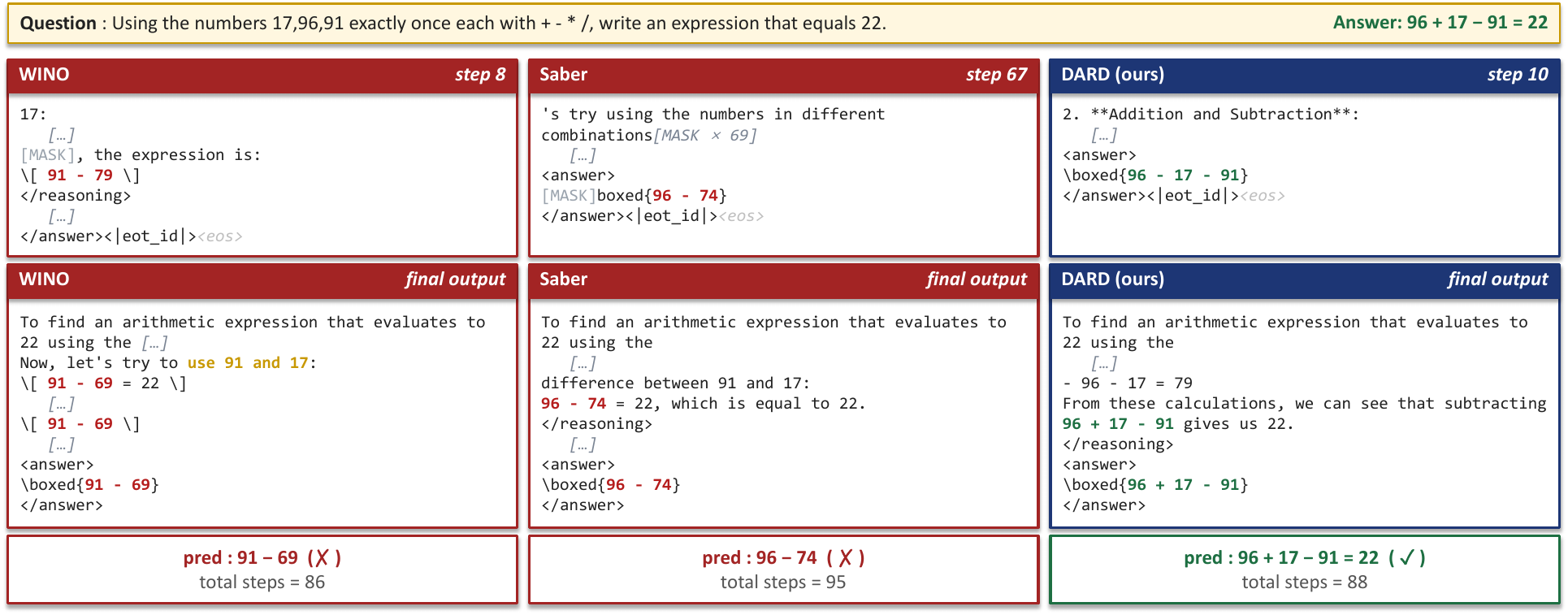}% or \columnwidth
  \caption{
Qualitative comparison of DARD with conventional revocable decoding methods.  }
  \label{fig:appendix_qualitative_1}
\end{figure*}

%% file: appendix/figures/appendix_qualitative_2.tex
\begin{figure*}[t!]
  \centering
  \includegraphics[width=\textwidth]{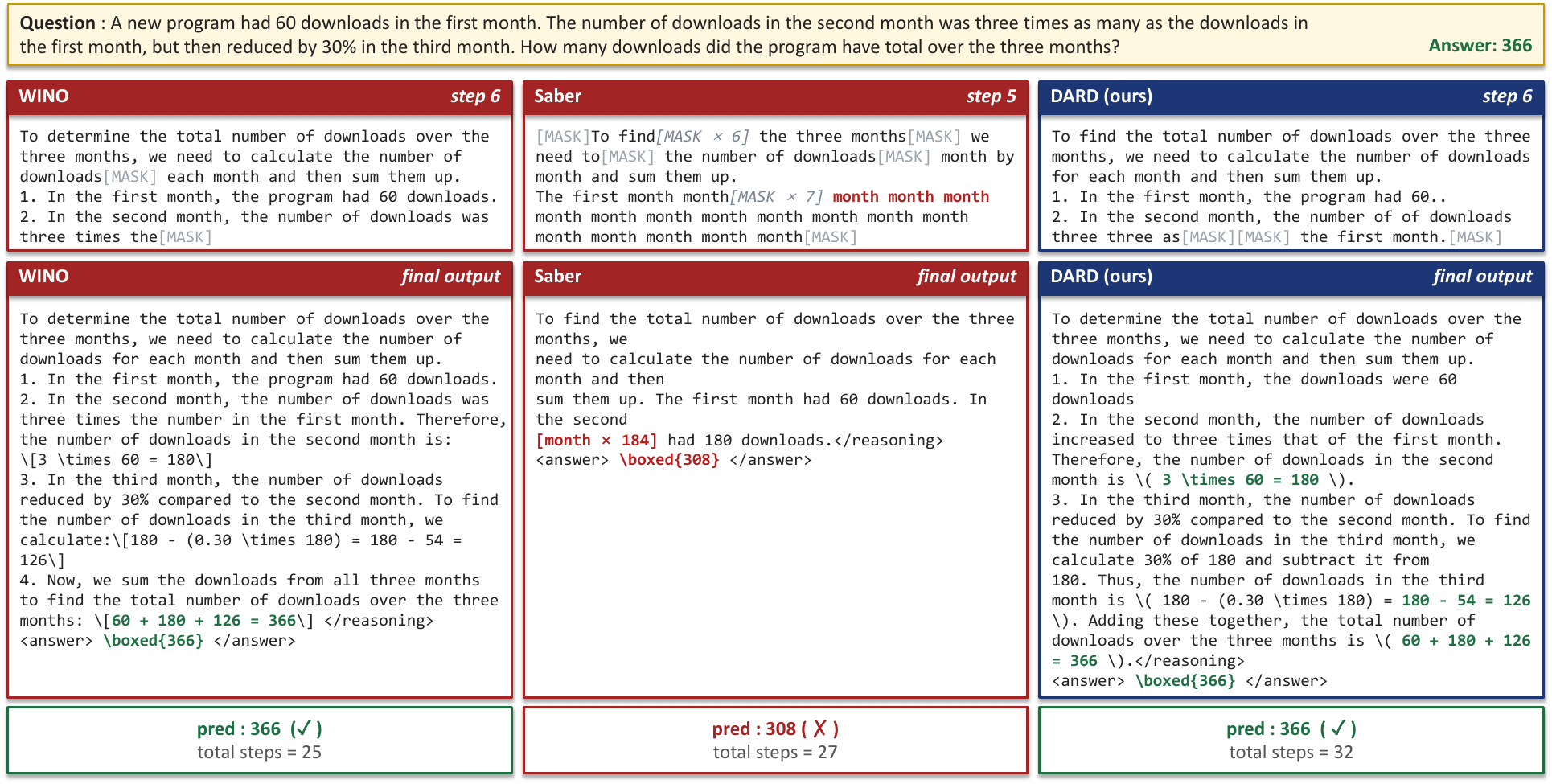}% or \columnwidth
  \caption{
Qualitative comparison of DARD with conventional revocable decoding methods.  }
  \label{fig:appendix_qualitative_2}
\end{figure*}

%% file: appendix/figures/appendix_qualitative_3.tex
\begin{figure*}[t!]
  \centering
  \includegraphics[width=\textwidth]{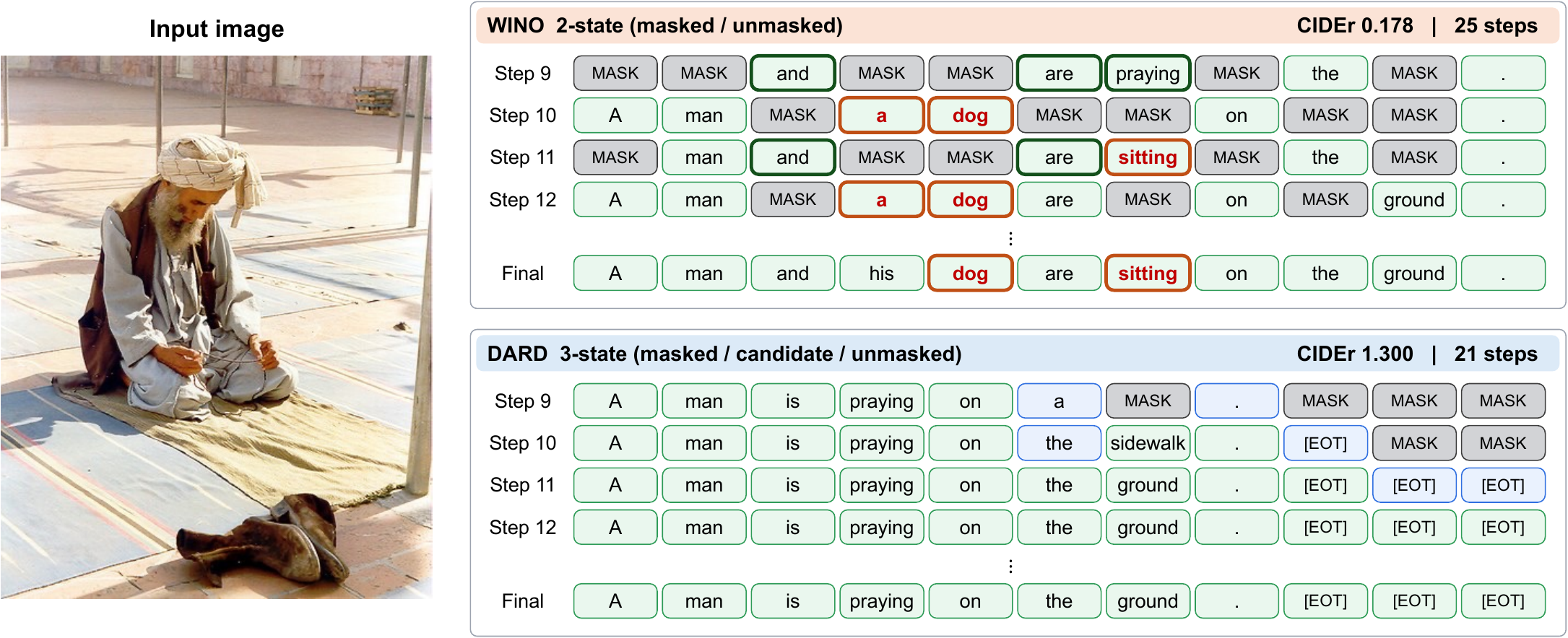}% or \columnwidth
  \caption{
Qualitative comparison between the decoding processes of conventional revocable decoding and DARD.
 }
  \label{fig:appendix_qualitative_3}
\end{figure*}

%% file: appendix/figures/appendix_qualitative_4.tex
\begin{figure*}[t!]
  \centering
  \includegraphics[width=\textwidth]{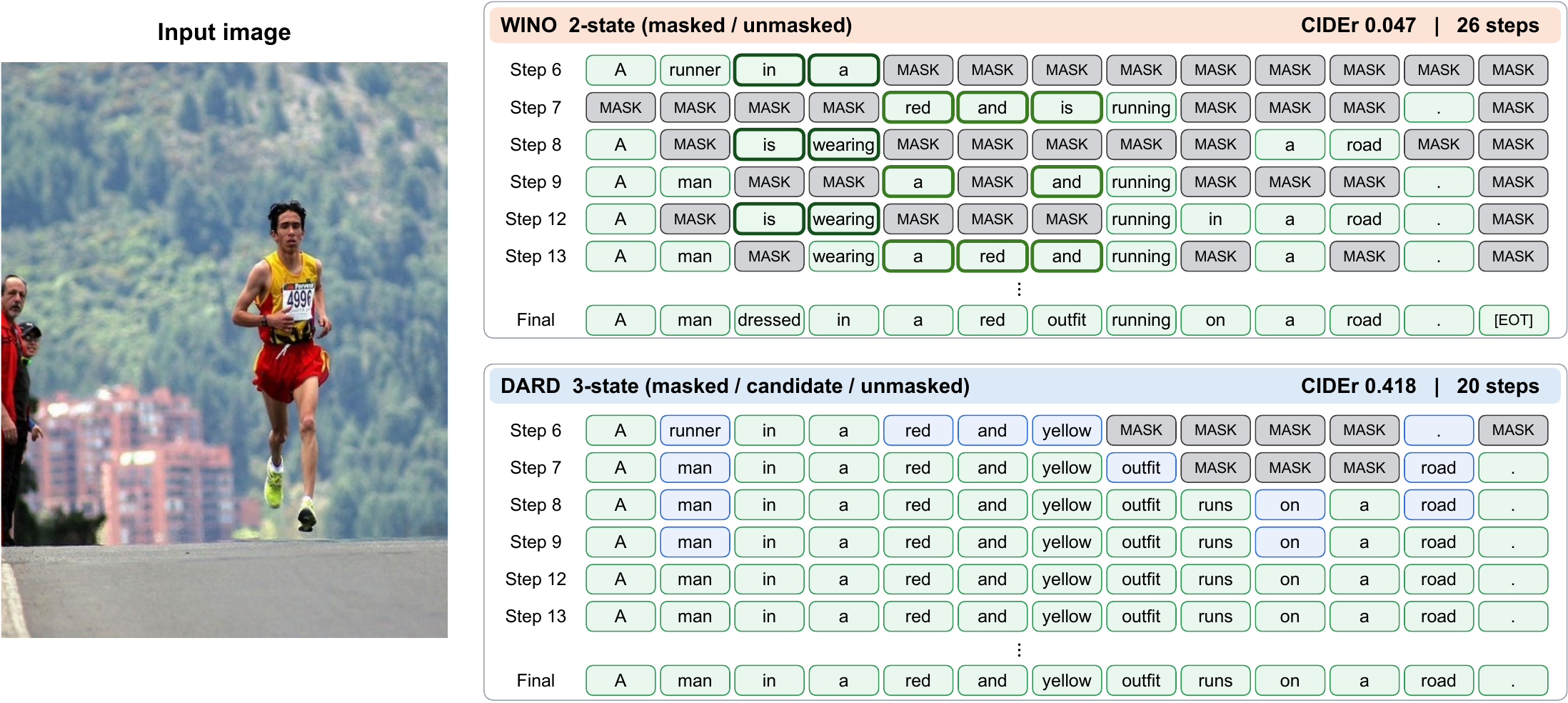}% or \columnwidth
  \caption{
Qualitative comparison between the decoding processes of conventional revocable decoding and DARD.  }
  \label{fig:appendix_qualitative_4}
\end{figure*}

%% file: appendix/tables/algorithm.tex
\begin{algorithm*}[t]
\caption{Dependency-Aware Revocable Decoding}
\label{alg:state_transition}
\begin{algorithmic}[1]
\Require Sequence $\mathbf{x}_0$, dLLM $p_\theta$, confidence thresholds
$\tau_{\mathrm{c}}, \tau_{\mathrm{u}}$, decay parameter $\lambda$, and prior constant $p_0$
\Ensure Decoded sequence $\mathbf{x}_T$

\State Initialize
$\mathcal{M}_0 \leftarrow \{0,\ldots,L-1\}$,
$\mathcal{C}_0 \leftarrow \emptyset$,
$\mathcal{U}_0 \leftarrow \emptyset$

\For{$t = 0,\ldots,T-1$}
    \State Initialize
    $\mathcal{M}_{t+1},\mathcal{C}_{t+1},\mathcal{U}_{t+1} \leftarrow \emptyset$
    \State Construct $\tilde{\mathbf{x}}_t = (\mathbf{x}_t;\mathbf{s}_t)$
    \State Build the attention mask $M$ for $\tilde{\mathbf{x}}_t$
    \State Obtain $\text{logit}_\theta(x_{t+1}^{i})$ and $\text{logit}_\theta(s_{t+1}^{{i}})$ for $i \in \mathcal{M}_t$ using $p_\theta(\mathbf{x}_{t+1} \mid \tilde{\mathbf{x}}_t;M)$ and $p_\theta(\mathbf{s}_{t+1} \mid \tilde{\mathbf{x}}_t;M)$, respectively.
    \State Obtain $c_t^{{i}}$ for $i \in \mathcal{C}_t \cup \mathcal{U}_t$ using $p_\theta(\mathbf{s}_{t+1} \mid \tilde{\mathbf{x}}_t;M)$.
    \State Initialize $\mathcal{P}_t \leftarrow \emptyset$ and
    $\mathcal{D}_t \leftarrow \emptyset$
    \For{$i \in \mathcal{C}_t$} \Comment{Update $\mathcal{C}_t$ states}
        \If{$\tau_{\mathrm{u}}<c_t^{{i}} $}
            \State Add $i$ to $\mathcal{U}_{t+1}$ and $\mathcal{P}_t$
        \ElsIf{$\tau_{\mathrm{c}} < c_t^{{i}} \leq \tau_{\mathrm{u}}$}
            \State Add $i$ to $\mathcal{C}_{t+1}$
        \Else
            \State Add $i$ to $\mathcal{M}_{t+1}$ and $\mathcal{D}_t$
        \EndIf
    \EndFor

    \For{$i \in \mathcal{M}_t$} \Comment{Decode $\mathcal{M}_t$ with logit mixing}
        \State Compute $w_t^i$, $P_t^i$ and $D_t^i$ using $\lambda$, and $p_0$
        \State Compute $\text{logit}_\theta^{\mathrm{mix}}$ using $w_t^i$
        \State Obtain confidence $c_t^i$ from $\text{logit}_\theta^{\mathrm{mix}}$
    \EndFor

    \For{$i \in \mathcal{M}_t \cup \mathcal{U}_t$}
    \Comment{Update $\mathcal{M}_t$ and $\mathcal{U}_t$ states}
        \If{$ \tau_{\mathrm{u}}<c_t^i$}
            \State Add $i$ to $\mathcal{U}_{t+1}$
        \ElsIf{$\tau_{\mathrm{c}} < c_t^i \leq \tau_{\mathrm{u}}$}
            \State Add $i$ to $\mathcal{C}_{t+1}$
        \Else
            \State Add $i$ to $\mathcal{M}_{t+1}$
        \EndIf
    \EndFor

    \State Fill positions in $\mathcal{C}_{t+1} \cup \mathcal{U}_{t+1}$ with
    decoded tokens, and keep positions in $\mathcal{M}_{t+1}$ as $\texttt{[Mask]}$
\EndFor

\State \Return $\mathbf{x}_T$
\end{algorithmic}
\end{algorithm*}